\documentclass{article}

\PassOptionsToPackage{numbers}{natbib}
\usepackage[preprint]{neurips_2026}
\usepackage{amsmath}
\usepackage{booktabs}
\usepackage{multirow}

\usepackage[utf8]{inputenc} % allow utf-8 input
\usepackage[T1]{fontenc}    % use 8-bit T1 fonts
\usepackage{hyperref}       % hyperlinks
\usepackage{url}            % simple URL typesetting
\usepackage{booktabs}       % professional-quality tables
\usepackage{amsfonts}       % blackboard math symbols
\usepackage{nicefrac}       % compact symbols for 1/2, etc.
\usepackage{microtype}      % microtypography
\usepackage{tabularx}
\usepackage{enumitem}
\usepackage{caption}
\usepackage{array}
\usepackage{xcolor} 
\usepackage{float}% colors
\usepackage{etoc}
\usepackage[most]{tcolorbox}
\tcbuselibrary{listings}
\usepackage{inconsolata} % nicer monospace font (optional)
\usepackage[disable]{todonotes}

\definecolor{promptgrey}{RGB}{245, 247, 250} 
\definecolor{promptgreyborder}{RGB}{200, 210, 220}

\definecolor{promptblue}{RGB}{235, 245, 255}
\definecolor{promptblueborder}{RGB}{180, 205, 230}

\definecolor{promptgreen}{RGB}{235, 250, 240}
\definecolor{promptgreenborder}{RGB}{180, 220, 190}

\definecolor{promptorange}{RGB}{255, 245, 230}
\definecolor{promptorangeborder}{RGB}{235, 200, 150}

\definecolor{promptpurple}{RGB}{245, 240, 255}
\definecolor{promptpurpleborder}{RGB}{210, 200, 235}

\newtcblisting{flexprompt}[3][]{
    colback=#2,        % background color
    colframe=#3,       % border color
    title=#1,          % title (optional)
    coltext=black,        % <-- main text color
    coltitle=black,       % <-- title text color
    fonttitle=\bfseries,
    fontupper=\ttfamily\small,
    listing only,
    boxrule=0.5pt,
    arc=4pt,
    boxsep=4pt,
    left=6pt, right=6pt, top=6pt, bottom=6pt,
    breakable,
    listing options={
        basicstyle=\ttfamily\small,
        breaklines=true,
        columns=fullflexible
    }
}

\definecolor{ameycol}{HTML}{D1E9FF}    % Soft Blue
\definecolor{ananyacol}{HTML}{D1FFD1}  % Soft Green
\definecolor{navincol}{HTML}{FFE4B5}   % Moccasin/Orange
\definecolor{ravicol}{HTML}{F3D1FF}    % Soft Purple

\newcommand{\amey}[1]{\todo[linecolor=blue, backgroundcolor=ameycol, bordercolor=blue]{Amey: #1}}
\newcommand{\ameyil}[1]{\todo[inline, backgroundcolor=ameycol, bordercolor=blue]{Amey: #1}}

\newcommand{\ananya}[1]{\todo[linecolor=green!60!black, backgroundcolor=ananyacol, bordercolor=green!60!black]{Ananya: #1}}

\newcommand{\navin}[1]{\todo[linecolor=orange!80!black, backgroundcolor=navincol, bordercolor=orange!80!black]{Navin: #1}}

\usepackage{enumitem} % Add to preamble

\title{\textsc{WinSyn}: An Automated Pipeline for Realistic Enterprise Question-Answering Evaluation}

\author{%
  Ananya Sutradhar\thanks{Equal contribution. Work done while at Microsoft Research} \\
  Microsoft Research\\
  India\\
  \texttt{sdhar.ananya@gmail.com} \\
  \And
  Amey Varhade\footnotemark[1] \\
  Microsoft Research\\
  India\\
  \texttt{ameyvarhade@gmail.com} \\
  \AND
  Ravishankar Krishnaswamy \\
  Microsoft Research\\
  India\\
  \texttt{rakri@microsoft.com} \\
  \And
  Navin Goyal \\
  Microsoft Research\\
  India\\
  \texttt{navingo@microsoft.com} \\
}

\usepackage{listings}
\begin{document}

\maketitle
\begin{abstract}
Enterprise settings provide a challenging environment for question-answering agents, which often rely on Retrieval-Augmented Generation, Deep Research (DR), and related techniques. Much of this challenge comes from the complexity of enterprise data: information is often spread across evolving and potentially conflicting emails, chat messages, documents, and other artifacts. Existing benchmarks typically have limited real-world complexity, short-form responses, and unnatural queries, so they often fail to capture the challenges of enterprise settings. In this work, we introduce an automated pipeline for generating synthetic datasets of emails reflecting realistic workplace scenarios, along with long- and short-form questions and gold answers grounded in the data.

Our method simulates long-running enterprise projects spanning several months and involving up to 25 interacting employees across multiple roles. The data emphasizes ambiguity, distributed information, and naturally occurring queries. To validate the pipeline, we evaluate few standard agentic baselines on our datasets using the latest frontier models. We find that aggregate scores averaged over all queries remain below 80\% for each dataset, indicating significant room for improvement.  
These findings suggest that more work remains to be done for enterprise deployment and underscore the importance of realistic, high-complexity evaluation data for developing stronger real-world enterprise DR systems.

\end{abstract}

\section{Introduction}

Enterprise question answering differs from standard document-grounded QA in a fundamental way: the answer is often not contained in any single artifact. A project decision may be introduced in an email, revised in a later discussion, justified in a planning document, and finally reflected in a ticket, launch note, or postmortem. In such settings, answering a query requires more than retrieving a relevant passage. A system must reconstruct an evolving organizational state from partial, distributed, and sometimes conflicting evidence.

This makes evaluation difficult. Real enterprise corpora are rarely available for research because they often contain private, proprietary, and compliance-sensitive information. As a result, recent work has increasingly relied on synthetic data for evaluating Retrieval-Augmented Generation (RAG), deep search, and agentic question-answering systems. However, generating useful synthetic enterprise data is itself challenging. A realistic workplace corpus is not just a collection of plausible documents: it must preserve temporal consistency, causal dependencies, role-specific communication patterns, implicit organizational context, and enough grounding structure to make reference answers auditable.

Existing synthetic enterprise benchmarks make important progress toward this goal, but they often involve trade-offs between realism, scalability, domain flexibility, and answer verifiability. Some pipelines rely on manually specified templates or workflows, which can limit adaptation to new domains; e.g., \cite{choubey2025herb, abaskohi2026drbench}. These limitations motivate a generation process that can maintain global coherence while still producing local artifact-level detail.

We introduce \textsc{WinSyn} (Workplace Interaction Synthesis), an automated pipeline for generating synthetic enterprise question-answering datasets from compact seed scenarios. Rather than simulating unconstrained agents end-to-end, \textsc{WinSyn} follows a top-down construction process. It first expands a seed document into company context, employees, project epics, task dependency graphs, and daily activity logs. These intermediate structures serve as a hidden source of truth for the simulated workplace. The pipeline then generates communication artifacts and grounded question--answer pairs from this scaffold.

A key design choice is to separate the internal generative scaffold from the evidence available at evaluation time. The pipeline uses structured intermediate representations to maintain temporal and causal consistency, but evaluated systems observe only the final enterprise artifacts. Thus, the benchmark tests whether systems can recover the relevant structure from fragmented workplace evidence, rather than relying on the privileged representations used during generation.

In this paper, we instantiate \textsc{WinSyn} with email corpora for simplicity, while designing the pipeline to be extensible to other workplace modalities such as chats, meetings, documents, tickets, and code repositories. We generate four enterprise QA datasets across different technical domains, including both synthetic seed scenarios and real public technical documents used as seeds. Each dataset contains a simulated multi-month workplace project involving more than twenty employees, together with questions, reference answers, and citation sets grounded in the generated data.

We evaluate representative agentic RAG and DR baselines on these datasets using retrieval and answer-quality metrics. Our results show that current systems still struggle with this setting, particularly when questions require synthesizing evidence across long-running projects rather than retrieving isolated facts. These findings suggest that temporally structured synthetic enterprise datasets can serve as useful stress tests for enterprise search and question-answering systems.

Our contributions are:
\begin{itemize}
    \item We present \textsc{WinSyn}, an automated top-down pipeline for generating temporally structured synthetic enterprise QA environments from compact seed scenarios.
    \item We introduce four generated enterprise email datasets with grounded questions, reference answers, and citation sets.
    \item We provide an evaluation of representative agentic QA systems across retrieval, grounding, and answer-quality dimensions, highlighting persistent gaps in current enterprise QA systems.
\end{itemize}

\textbf{Organization.} The next section first discusses relevant properties of enterprise data followed by our pipeline. Sec.~\ref{sec:related_work} discusses related works and Sec.~\ref{sec:experiements} details the results of our experiments. We conclude in Sec.~\ref{sec:conclusion}.

\section{\textsc{WinSyn:} Automated Pipeline for Enterprise Data Generation}
In this section we outline how our pipeline works. We begin by discussing characteristics of enterprise data that motivate our pipeline design. 

\subsection{Enterprise data.} \label{subsec:enteprise_data} To illustrate what we mean by enterprise data, consider a team of twenty people working on a project over three months. The team plans, executes, pivots, and re-executes. Much of this activity gets recorded in artifacts: emails, chat messages, documents, meeting transcripts, pull requests, and more. Together these form a dataset (of enterprise data).

In addition to the enterprise data, our dataset also consists of questions and gold answers (QAs) which use the enterprise data as their basis. These will be used to evaluate and/or train AI agents. How do we guarantee that the gold answers are correct given the enterprise data?

To answer complex queries over such a dataset: why was a decision made, what caused an incident, who knew what and when---one must reconstruct causal chains that are distributed across many artifacts. These artifacts evolve over time and can be mutually conflicting. Thus, linking them into a coherent picture is one basic challenge. Others include judging the authoritativeness or currency of a piece of information, and navigating an organization's internal jargon and culture.

\paragraph{The structure of enterprise data.} A realistic enterprise dataset has a rich structure spanning multiple scales and dimensions which are closer in character to a story or play than to a document collection.

There are characters with personalities, goals, and actions; events cause new events; there is a distributed world-state that evolves as actions are taken. This causal structure operates at multiple scales:
\begin{itemize}[leftmargin=*]
\item{\textit{Micro.}} A customer sends a support email about a bug → a support agent flags it in a Teams channel → a developer opens a ticket. The data schema does not natively link these artifacts; reconstructing those links is part of the challenge.

\item{\textit{Macro.}} A CEO declares a company-wide OKR → departments spin up projects, generating trails of meetings and specs → a product launches → incident reports and retrospectives are filed.
\end{itemize}

Many intermediate scales exist between these poles. A quarterly report compresses months of activity; a meeting transcript records events in real time. A single decision typically spans artifact types: a problem surfaces in an email, options are debated in a meeting, a recommendation is written up, the decision recorded in minutes, and implementation tracked in tickets.

Like a story, the dataset must be internally consistent and free of plot holes; e.g., \cite{ahuja2025findingflawedfictionsevaluating}. It has directionality arising from people pursuing their goals. Later artifacts can contradict earlier ones (a design decision reversed after implementation work, for instance), but such reversals must themselves be recorded and explained; they are part of the consistency, not violations of it.

The dataset is organized around entities (people, projects, products, clients, dates) whose relationships are asserted, assumed, or implied across documents. Running through it are softer themes: company culture, formal and informal social networks, collective sentiment (discovery mode and crisis mode). Much of this context is never made explicit; artifacts presuppose shared history, vocabulary, and relationships that a reader must infer.
Finally, the dataset is not closed. Relevant context lives outside it, in people's heads, in external documents, in conversations that were never recorded.

Enterprise data is therefore best understood as a distributed, multi-agent, temporally extended discourse, not a collection of documents.

\paragraph{Other dimensions of realism.} The discussion so far has emphasized logical and causal structure. Realism has many other dimensions too. It's not possible to be exhaustive; we list only some: 

\textit{Statistical realism.} Word-count distributions, sentence lengths, response-time patterns, and the diversity of personas, styles, tones, and sentiments must all reflect real enterprise communication.

\textit{Pragmatic realism.} A ``reply all'' with ``sounds good'' carries implicit commitment and closes a thread. Much of the real content of company communication lives in pragmatic inference, not literal content. Rejected options, abandoned threads, and unanswered emails are causally meaningful---they mark paths not taken.

\iffalse
Epistemic realism. Artifacts encode what was known at the time. The dataset as a whole traces the evolution of collective knowledge: when did the company learn X? Who knew it first? Who was out of the loop when they should not have been?
\fi

\textit{Survival bias.} Not all artifacts survive equally. Email threads persist; hallway conversations do not. A realistic dataset must account for this systematic gap and simulate the traces that ephemeral communication leaves behind in other artifacts.

\subsection{Our pipeline}
Designing a pipeline to automatically construct enterprise datasets with a rich structure as mentioned above is the problem we study in this paper. 

\textbf{Challenges.} Perhaps the first idea that comes to mind for the construction of such a dataset is to run a simulation of agents: create employees with personas, give them goals, and let them interact to achieve those goals. We do not take this approach as it is hard to ensure that these interactions remain coherent, achieve goals and do not go off on tangents. Moreover, the problem of constructing QAs for such a dataset runs into a circularity issue: how do we know the gold answer to a given question? 
Apart from addressing these challenges, our approach below is designed to deal with various other challenges that arise: keeping the whole dataset realistic and consistent, context size limitations, subtle instruction following failures, and hallucinations, even in the best frontier models. 

For simplicity we focus on emails in this paper. Our approach generalizes to other modalities, however more work in terms of adding more stages to the pipeline is needed. As mentioned earlier, real enterprise data has a complex structure; our pipeline approaches it in many aspects but not all: we emphasize the causal structure and consistency, and do not explicitly account for statistical and pragmatic realism. 

\textbf{Multi-stage pipeline.} Our hierarchical, multi-stage pipeline starts with a seed document which is a short narrative summarizing the main events that happened during the period covered by the dataset. This document could be company blog posts, internal newsletters, a changelog, a quarterly report etc. It could be a real-world document or could be constructed synthetically with some initial input from the user (size of the company, industry sector, project, events, etc.). The idea is to progressively expand this narrative into enterprise data by filling in more details at each expansion stage. Each stage after generation runs a number of validation checks, both programmatic and LLM-based. This expansion roughly follows the multiscale structure of enterprise data alluded to above: higher-level events and actions are synthesized first, and their lower-level counterparts after them. In fact, we use terminology similar to the popular \emph{Agile Methodology} \cite{beck2001manifesto} for some of the stages. Let's briefly recall what it is.

\textbf{Agile Methodology} is a popular iterative, flexible project management approach that breaks large projects into small, manageable pieces of work. \emph{Epics} are large bodies of work broken into smaller \emph{Tasks}. Epics typically span weeks to months, while Tasks last a week or two. We use two hierarchy levels for simplicity, though larger projects may use more. Rather than generating independent Epics, we construct a Directed Acyclic Graph (DAG) capturing dependencies between them and record the artifacts handed over. Similarly, each Epic is organized as a Task DAG. We remark that our technique is not tied to Agile Methodology; instead, it relies on the fact that any complex workflow must be hierarchical to deal with the complexity.

The multi-stage pipeline design mirrors real organizational workflows and ensures that all generated
content remains coherent and consistent across temporal, organizational, and informational dimensions as the generated data at each stage is faithful to the previous stage's data. The circularity problem is avoided by keeping a concise source of truth in which all answers are grounded as explained below. It also allows us to keep the context size small: since we know the hierarchical structure of the data, when generating any particular piece of information, say a task, we know the most pertinent information (e.g., the Epic containing that Task, related other epics, predecessor tasks) which we can provide in context instead of having to provide the whole dataset from the previous stage. Note that at test time, the agent does not have the benefit of this information and must work with emails to understand the structure of the data. 

Now we list the steps for dataset creation:

\begin{enumerate}[leftmargin=*]
\item{\textbf{Company information.}} In this step, company name, industry sector, size are first generated, followed by employees, their roles in the company, reporting structure, personal and technical background of the employees. 
\item{\textbf{Epics.}} We create 5-10 Epics along with a dependency directed acyclic graph (DAG) between them. 
For each Epic the generation includes a title, a scoped description, assigned employees, a business-day timeline, inter-epic dependencies (which epics need to complete for the current epic to start, which other epics depend on the current epic; what are the artifacts to be received and handed off). (An example Epic is Fig. \ref{fig:artifactepic})

\item{\textbf{Tasks.}} For each Epic 5-10 Tasks are generated along with a DAG. Each task has associated information similar to epics. Refer to an example task in Fig. \ref{fig:artifacttask}.

After task generation we depart from the Agile methodology.

\item{\textbf{Daily Diaries.}} These are obtained by expanding task descriptions into a log of daily activity. For each Task and each day in its timeline, we emit a structured diary entry comprising fields for work summary, per-employee detailed activities, progress notes, blockers (with
reporter), resolutions (with resolver), collaboration interactions (with participant lists).

Daily diaries (Such as Fig. \ref{fig:artifactdiary}) form the backbone of the full dataset and serves as a concise source of truth in which both the QAs and the emails and other final communication artifacts are grounded. 
\item{\textbf{QA.}} As mentioned earlier, 
the main idea for ensuring that the answers in our dataset are indeed correct answers is to control the context size when generating the answers. This can be done because we know the hierarchical structure of the data. Some examples include Fig. \ref{fig:outputdailybriefingqa}, \ref{fig:outputprojectstatusqa}  \ref{fig:outputmultihopqa} and \ref{fig:outputrecapqa}

\item{\textbf{Emails.}} Emails (and other artifact types e.g. chats, if included) are generated from the daily logs. The main idea here is that emails fill in lower-level details that do not alter the answers to any questions in QA. \navin{define notation QA somewhere earlier} Emails add significantly more details to daily logs. As mentioned previously, an agent working with emails to answer queries will have a significantly harder task at hand than our pipeline does; it needs to understand the full dataset (unlike our pipeline, it is not provided with the underlying recursive structure and just the right context when answering a query). 
\end{enumerate}

Other parts of the pipeline include sentence-level attribution of answers into emails, addition of distractors to emails for added realism, and extensive validation checks and reflection-refinement loops. While conceptually secondary,  validation checks and reflection-refinement loops are an essential part of the pipeline as even the best frontier LLMs invariably make subtle and not-so-subtle errors in instruction following in our use case despite extensive revision of instructions; a substantial amount of effort went in their design. One example of such errors is \textit{future leaks}: description of an Epic sometimes erroneously refers to something that happens in the future; our pipeline fixes such issues. Find a sample email thread at Fig. \ref{fig:outputemailthread}
A detailed discussion of our pipeline is in Appendix \ref{app:B}. The Fig. \ref{fig:pipeline} shows the overall flow of our pipeline from the seed document to the final emails and QA which make up the dataset.

\begin{figure}[t]
    \centering
    \includegraphics[width=\linewidth]{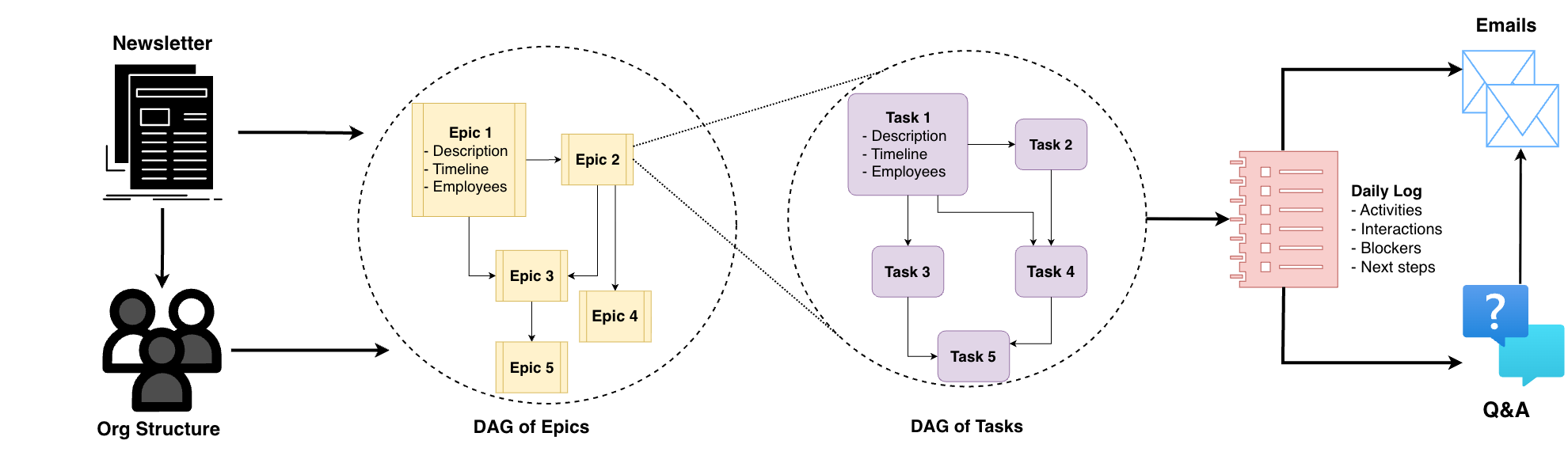}
    \caption{The pipeline starts with seed document (e.g. newsletter) and creates an organizational structure from its content. Both of them are used to create a DAG of Epics each of which is expanded into a DAG of Tasks. Each Task involves daily log for its duration and then the Emails are generated followed by the question-answers and the attributions of emails}
    \label{fig:pipeline}
\end{figure}

\section{Related Work}
\label{sec:related_work}
\paragraph{Story writing.} As noted earlier, enterprise data shares some similarities with stories while also having many differences. A brief note on how our pipeline design relates to story writing may be illuminating. Most fiction authors fall somewhere on a spectrum defined by how much they plan before they start writing the story proper. The extreme ends of this spectrum are occupied by 
the \emph{pantsers/gardners} and \emph{plotters/architects}; e.g., \cite{king2000onwriting}. The former start with a premise and develop it without knowing what the outcome would be, whereas the latter plan everything ahead of actual writing. The agent simulation method mentioned before represents the pantsers style. Our approach in the present paper is closer to the plotters style. The reason for choosing this, as mentioned earlier, is to keep the story coherent as well as have a firm grasp on the ground truth. A hybrid between the two methods may lead to even more realistic enterprise data but we leave this exploration to the future work.
There's also significant literature on AI-generated fiction, e.g., \cite{teleki-etal-2025-survey}. To our knowledge, fiction generation frameworks are not directly adaptable for synthetic enterprise data generation. 

\textbf{Existing benchmarks.}
Starting with early datasets like HotpotQA \cite{yang2018hotpotqa},
a large number of new benchmarks have been constructed for information retrieval tasks and also related tasks where information retrieval serves as an important backbone. We can only give a very small sample: 
for RAG over research papers \cite{asai2024openscholarsynthesizingscientificliterature}, for deep research \cite{gupta2026deepsearchqabridgingcomprehensivenessgap}, and for other related tasks OfficeQA Pro~\cite{databricks2025officeqa}, DRACO \cite{zhong2026dracocrossdomainbenchmarkdeep}, APEX-Agents \cite{vidgen2026apexagents}, PRBench \cite{akyurek2025prbench}, TheAgentCompany \cite{xu2025agentcompany}.
Among the synthetic enterprise datasets, closest to the present work are DRBench and HERB:

HERB~\cite{choubey2025herb} benchmarks RAG over 39,190 synthetic enterprise artifacts (Slack, documents, GitHub PRs, meeting transcripts) spanning 30 software products. Its query-first design: human experts define queries before generating supporting evidence via LLM-simulated workflows which guarantees traceable ground truth and natural multi-hop questions grounded in realistic enterprise processes. The questions require aggregating evidence across multiple heterogeneous sources, making the benchmark a meaningful test of retrieval completeness. However, the pipeline is domain-specific: Its query templates and workflow specifications are hand-engineered for the software lifecycle, requiring substantial expert re-design to apply to other domains.

DRBench~\cite{abaskohi2026drbench} benchmarks AI agents on open-ended deep research tasks in enterprise settings. Includes queries such as "How can personalization drive sales and what strategies can Lee's Market use?" requiring multi-step synthesis across private enterprise data and the public web, evaluated on insight recall, factuality, and report quality. Each task provides ~20 files spanning DOCX, XLSX, Mattermost chat logs, and emails, with ground-truth insights distributed across all source types. However, with only ~0.5 external facts per task on average, the primary challenge is reasoning over internal enterprise data rather than web retrieval. Critically, the enterprise files are static and non-conflicting: chat logs and emails serve as containers for pre-planted facts rather than evolving dialogues, making the core challenge one of multi-source fact retrieval rather than temporal reasoning. This contrasts with HERB, where versioned documents, feature reversals, and iterative Slack discussions require the model to track which version of a fact is current. Like HERB, DRBench is a fixed benchmark not designed for adaptation to new enterprise contexts.

\iffalse
\textbf{PRBench}~\cite{akyurek2025prbench} focuses on high-stakes professional reasoning in law and finance. It comprises 1,100 expert-authored tasks, each accompanied by detailed rubrics with 10--30 weighted criteria. Tasks are inspired by real-world workflows and require nuanced, economically consequential decision-making. Despite the use of frontier models, performance remains well below reliability thresholds, highlighting the difficulty of domain-specific reasoning.

\textbf{OfficeQA} and its successor \textbf{OfficeQA Pro}~\cite{databricks2025officeqa} evaluate grounded reasoning over large-scale, document-heavy corpora. The benchmark uses 89,000 pages of U.S. Treasury Bulletins as a public proxy for enterprise documents. Tasks involve multi-hop reasoning over scanned PDFs, complex tables, and temporal data. Even with structured document parsing, leading agents achieve less than 70\% accuracy, underscoring the challenges of document understanding and retrieval.

\textbf{TheAgentCompany}~\cite{xu2025agentcompany} simulates a small software company environment to evaluate LLM agents on 175 real-world workplace tasks across domains such as HR, finance, and engineering. Agents interact with tools like GitLab and Rocket.Chat, and are evaluated using checkpoint-based rubrics. Results show that even top-performing agents autonomously complete only $\sim$30\% of tasks, with long-horizon planning and collaboration remaining key bottlenecks.

\fi

Other somewhat closely related work includes CRMArena-Pro~\cite{huang2025crmarena}, and DataMorgana~\cite{filice-etal-2025-generating}.

\iffalse 
\textbf{Mind2Web 2}~\cite{gou2025mind2web} evaluates agentic web search capabilities across 300 Tasks on live websites. Though not enterprise-specific, it informs evaluation methodologies (e.g., LLM-as-a-judge) now adopted in enterprise benchmarks like DRBench and PRBench.

Complementary efforts include \textbf{GDPval}~\cite{patwardhan2025gdpval}, which evaluates economically valuable tasks in a closed-world setting, and \textbf{ProBench}~\cite{yang2025probench}, which targets multimodal expert-level queries across diverse domains. \textbf{ProfBench}~\cite{wang2025profbench} further expands this space with over 7,000 rubric-annotated tasks in domains such as physics, chemistry, and finance.
\fi

%In the domain of synthetic data generation, \textbf{SDGym}~\cite{sdgym2024} and the evaluation framework by \textbf{Sidorenko et al.}~\citep{sidorenko2024synthetic} provide standardized metrics for assessing fidelity and privacy in synthetic tabular data. These tools are foundational for constructing synthetic enterprise corpora used in benchmarks like HERB and DRBench.

\iffalse
Together, these benchmarks reflect a shift toward evaluating AI systems in realistic, high-stakes, and heterogeneous enterprise environments. They also highlight the growing role of synthetic data pipelines in constructing scalable, privacy-preserving evaluation corpora. Despite progress, current models exhibit significant limitations in retrieval, reasoning, and domain-specific judgment, motivating continued research in enterprise-grade AI evaluation.
\fi

\section{Experiments}
\label{sec:experiements}

% Generation is parallelised across the topological levels of
% $\mathcal{G}_E$ and, within each epic $e \in V_E$, across up to five
% tasks $\tau \in V_T^{(e)}$. For a given task-day $(\tau, d)$, meetings
% $\mathcal{M}_{\tau,d}$ and chats $\mathcal{C}_{\tau,d}$ are produced
% concurrently, while email threads $\mathcal{L}^{(\tau)}$ are generated
% sequentially.

% All intermediate artifacts---including initial epics, per-iteration
% refinement outputs, task graphs $\mathcal{G}_T^{(e)}$, diary entries
% $\{D_{\tau,d}\}$, communication artifacts
% $\{\mathcal{L}^{(\tau)}, \mathcal{M}_{\tau,d}, \mathcal{C}_{\tau,d}\}$,
% and validation reports---are checkpointed to persistent storage.
% Re-execution resumes from the latest completed stage, avoiding
% redundant LLM calls.

% A typical end-to-end run yields approximately $7$ epics,
% $40$ tasks, $1{,}200$ diary entries, $250$ emails, $85$ meetings,
% $120$ chat messages, and $12$ grounded question--answer pairs over a
% $60$-day window. This corresponds to a sentence set
% $\mathcal{S}$ comprising several hundred indexed records and the
% resulting attribution map $\alpha$.

\subsection{Datasets}

We consider four different datasets, each representing a distinct team within a company. Each dataset consists of the workplace simulation of the respective team executing a particular project. The four datasets we have generated include teams for (i) Cloud DevOps Platform team, (ii) Data Platform Engineering Team, (iii) Stripe Billing, and (iv) VS Code Changelog. The datasets vary by domain and source: Cloud DevOps Platform and Data Platform Engineering use LLM-generated synthetic enterprise newsletters, while Stripe Billing and VS Code Changelog are based on real public technical documents from the Stripe engineering blog and official VS Code release notes, respectively. These source documents are transformed into long-running workplace simulations through our automated generation pipeline. 
% Each dataset consists of 127--181 emails exchanged among 20--25 simulated employees, accompanied by 22--28 evaluation questions with gold reference answers and citation sets.

\ananya{more details/differences}
\ananya{introduce the question types properly}

Our dataset covers a diverse set of enterprise query types that reflect realistic workplace information needs. These range from short, fact-centric queries such as ownership attribution and action item retrieval, to more complex, long-form queries requiring synthesis across multiple sources, including project status reports, retrospectives, and multi-domain summaries.  The query set is designed to capture some of the key challenges in enterprise search, including multi-hop reasoning, attribution of responsibility across collaborators, temporal reasoning over evolving workstreams, and aggregation of distributed information. In addition, we include structured, high-level queries such as daily briefings and project status reports, which require the system to generate coherent, multi-section responses grounded in heterogeneous evidence.

A complete taxonomy of query types, along with representative examples and expected answer lengths, is provided in Table ~\ref{tab:question-types}. The Table \ref{tab:datasets} shows the dataset, original and overall statistics.

\ananya{mention how the synthetic newsletter generation}
\ananya{add qualitative charact. of the newsletters}
\ananya{complexity of the newsletters}
\ananya{add url}

\begin{table}[ht]
\centering
\small
\begin{tabular}{p{2cm} p{2cm} p{3.5cm} c r r r}
\toprule
\textbf{Dataset} & \textbf{Seed Doc.} & \textbf{Content} & \textbf{Emails} & \textbf{Emp.} & \textbf{Days} & \textbf{Questions} \\
\midrule
Stripe & Stripe engineering blog post & Real-time billing analytics with Apache Flink and Pinot & 127 & 22 & 117 & 28 \\
VS Code Changelog & Official VS Code Oct. 2025 release notes & IDE changelog covering agents, MCP, terminal, and other features & 142 & 23 & 100 & 22 \\
Data Platform Eng. & Synthetic & Lakehouse migration, streaming, data quality, and costs & 181 & 25 & 110 & 25 \\
Cloud DevOps & Synthetic & Developer portal, OpenTofu migration, Kubernetes, CI/CD, and cloud costs & 151 & 24 & 106 & 25 \\
\bottomrule
\end{tabular}
\vspace{2mm}
\caption{Dataset sources, content, and statistics.}
\label{tab:datasets}
\vspace{-2mm}
\end{table}

\subsection{Evaluations}
We use our dataset to evaluate two systems. These systems include the ReAct Agent defined in ~\cite{choubey2025herb} and an adaptation of the open source Deep Research (DR) Agent Onyx ~\cite{onyx2024}. A detailed example of a question answer pair and the citations outlining the quality and depth in the dataset is added in Appendix \ref{sec:detailedexample}.

\textbf{RAG (ReAct).}
\ananya{compare the gpt5.4 vs gpt4o}
We adopt this configuration as our primary baseline following Choubey et al.~\cite{choubey2025herb}, who show that a hybrid BM25 + dense retriever, when paired with a ReAct agent, is the best-performing system in their benchmark study, among all RAG approaches evaluated in HERB, including graph-based methods such as GraphRAG and HippoRAG-2, this agentic hybrid configuration achieves the highest overall performance, making it a natural reference point.

The system is built around ReAct~\cite{yao2023react}, a framework that interleaves reasoning and acting in an iterative loop. At each step, the agent first generates a natural-language \emph{thought} describing what information is still needed, then selects and invokes a tool (e.g., hybrid document retrieval, employee lookup), and observes the result before proceeding. This think--act--observe cycle enables the agent to adaptively refine its search strategy mid-trajectory, making it well-suited for the multi-hop, cross-source queries in HERB.

% The retriever is a hybrid BM25 + dense vector ensemble implemented using LlamaIndex, retrieving up to $k = 60$ document chunks per query. Documents are split into chunks of up to 256 tokens using a sentence-aware splitter (\texttt{LlamaIndex SentenceSplitter}), with a 32-token overlap between consecutive chunks to preserve local context across boundaries. The agent iterates for up to 10 reasoning steps, invoking the retriever and auxiliary metadata tools (e.g., employee lookup, role search) as needed.

\textbf{Adaptation of Onyx Deep Research Agent.}
\textbf{Onyx}~\cite{onyx2024} is a leading agentic RAG platform. It includes a Deep Research (DR) Agent. 
Our agent is a sandboxed adaptation of Onyx's Deep Research loop, restricted to a per-question visible email corpus. It first issues a planning call that returns 2--5 sub-questions and 1--4 seed tool calls, then enters a multi cycle \emph{act $\to$ observe $\to$ reflect} loop (up to ten cycles) before synthesizing a final answer. On each cycle the LLM may emit a single action or a batch of 2--4 complementary actions drawn from six local tools, \texttt{hybrid\_search}, \texttt{cosine\_search}, \texttt{bm25\_search}, \texttt{lexical\_search}, \texttt{grep\_search}, and \texttt{find\_files}, all of which are scoped to the visible corpus; web search and all non-local tools are disabled. 

\iffalse
\begin{table}[h]
\centering
\
\label{tab:config}
\begin{tabular}{lll}
\toprule
\textbf{Parameter} & \textbf{HERB RAG} & \textbf{Onyx DR} \\
\midrule
Generator model      & GPT-5.4                & GPT-5.4 \\
Embedding model      & \texttt{text-emb-ada-002}         & \texttt{text-emb-ada-002} \\
Retrieval mode       & Hybrid BM25 + vector (LlamaIndex)   & Vespa-based hybrid \\
\texttt{top\_k}      & 60                                  & 50 \\
Chunk size / overlap & 256 tokens / 32 tokens              & 1{,}600 chars / 200 chars \\
Agent iterations     & Up to 10 ReAct steps                & Up to 10 DR cycles \\

\bottomrule
\end{tabular}
\vspace{2mm}
\caption{System configurations.}
\end{table}
\fi

We use \texttt{gpt-5.4} as the underlying language model and \texttt{text-emb-ada-002} for embeddings.\footnote{\url{https://platform.openai.com/docs/models}, \url{https://platform.openai.com/docs/guides/embeddings}}

% \vspace{-6mm}

\subsection{Evaluation Metrics}
All systems are evaluated using an LLM-as-a-judge protocol with \texttt{gpt-5.4} as the evaluator. 
For each query, the judge is provided with the \textit{question}, the \textit{reference answer}, 
the \textit{candidate answer}, and (where applicable) the \textit{retrieved documents} and 
\textit{gold citation documents}. The judge evaluates system outputs across multiple dimensions 
using structured prompts (see Appendix~\ref{app:eval-prompts}). 

We report six primary metrics \textit{completeness}, \textit{soundness}, 
\textit{relevance}, \textit{faithfulness}, \textit{readability}, and \textit{retrieval accuracy}, all summarized in Table~\ref{tab:metrics}. Each metric is scored in the range $[0,1]$, except for the holistic \textit{overall score}, 
which is reported on a Likert scale from 1 to 5.

A weighted composite \textit{final score} is computed as:
\begin{equation}
\texttt{Final} = 0.30 \cdot \texttt{Comp.} + 0.30 \cdot \texttt{Sound.} + 0.15 \cdot \texttt{Relev.} + 0.15 \cdot \texttt{Faith.} + 0.10 \cdot \texttt{Read.}
\label{eq:final-score}
\end{equation}

The final score or ranking is to be computed based on the importance of metrics for the use case, the weights chosen are such that we emphasize on the completeness and soundness as we want to focus on the utility of the answers to the asker. Metrics such as relevance, faithfulness and retrieval are intermediate metrics to diagnose the retrieval component of the search system. We’ve come up with weights in Eq. \eqref{eq:final-score} based on our own understanding of the relative importance of the individual metrics. We believe that an unweighted average does not justify an overall score. This has been inspired by similar strategies employed by various RAG implementations (such as \cite{databricksRAG}).

In addition to judge-based metrics, we report standard information retrieval metrics to evaluate retrieval quality independently of generation. These include \textit{Mean Reciprocal Rank (MRR)}, \textit{Hit Rate@$k$}, \textit{Recall@$k$}
, and \textit{Citation Coverage Recall}, defined as the fraction of gold 
citation documents that appear anywhere in the retrieved set.

\begin{table}[t]
\centering
\small
\setlength{\tabcolsep}{3pt}
\begin{tabular}{p{2.6cm} p{5.0cm} c p{3.4cm}}
\toprule
\textbf{Metric} & \textbf{Description} & \textbf{Scale} & \textbf{Inputs to Judge} \\
\midrule
Completeness & Coverage of key information from the reference answer & $[0,1]$ & Q, Ref, Cand \\
Soundness & Factual accuracy; absence of contradictions vs.\ the reference & $[0,1]$ & Q, Ref, Cand \\
Relevance & Answer focus; absence of off-topic content & $[0,1]$ & Q, Ref, Cand \\
Faithfulness & Grounding of claims in retrieved evidence & $[0,1]$ & Q, Cand, Retrieved \\
Readability & Structural clarity and formatting & $[0,1]$ & Q, Cand \\
Retrieval Acc. & Quality of retrieved documents against gold citations & $[0,1]$ & Q, Retrieved, Gold \\
Overall Score & Holistic Likert-style quality judgment & $[1,5]$ & Q, Ref, Cand, Retrieved, Gold, Sub-scores \\
\bottomrule
\vspace{2mm}
\end{tabular}
\caption{LLM judge metrics. The abbreviation references are as Q: question; Ref: reference answer; Cand: candidate answer; Retrieved: retrieved chunks; Gold: gold citation documents; Sub-scores: scores from the preceding six metrics.}
\label{tab:metrics}
\vspace{-5mm}
\end{table}
% \vspace{-25mm}

\subsection{Results}
\label{sec:references}

We report the scores for the 4 datasets
Table~\ref{tab:dataset-results} reports per-dataset scores. Final scores are relatively close across both systems on all four datasets (within 0.11), although ReAct consistently achieves higher final scores across all datasets. In contrast, Onyx often attains higher relevance and soundness scores, indicating a tendency toward more precise but less comprehensive responses.
MRR values remain modest overall, suggesting that retrieving the most relevant document at rank 1 remains challenging for both systems. However, Onyx achieves noticeably stronger MRR on some datasets (e.g., Data Platform Eng.), indicating more effective early-rank retrieval despite lower end-to-end answer quality.

\begin{table}[H]
\centering
\small
\setlength{\tabcolsep}{4pt}
\begin{tabular}{l l c c c c c c c}
\toprule
\textbf{Dataset} & \textbf{System} & \textbf{Final} & \textbf{Comp.} & \textbf{Sound.} & \textbf{Relev.} & \textbf{Faith.} & \textbf{Overall} & \textbf{MRR} \\
\midrule
Stripe & ReAct & 0.755 & 0.607 & 0.777 & 0.704 & 0.928 & 2.845 & 0.475 \\
 & Onyx & 0.710 & 0.431 & 0.808 & 0.746 & 0.912 & 2.964 & 0.333 \\

VS Code Changelog & ReAct & 0.707 & 0.517 & 0.752 & 0.637 & 0.909 & 2.788 & 0.392 \\
 & Onyx & 0.694 & 0.414 & 0.803 & 0.712 & 0.892 & 2.636 & 0.400 \\
 
Cloud DevOps & ReAct & 0.792 & 0.672 & 0.883 & 0.663 & 0.898 & 3.120 & 0.430 \\
 & Onyx & 0.686 & 0.450 & 0.747 & 0.706 & 0.879 & 2.720 & 0.459 \\

Data Platform Eng. & ReAct & 0.752 & 0.640 & 0.806 & 0.587 & 0.933 & 2.960 & 0.357 \\
 & Onyx & 0.703 & 0.452 & 0.784 & 0.709 & 0.915 & 2.800 & 0.596 \\

\bottomrule
\end{tabular}
\vspace{1.5mm}
\caption{Performance comparison across datasets for the ReAct and Onyx Agents}
\label{tab:dataset-results}
\end{table}

To understand whether failure patterns vary by task structure, Table~\ref{tab:answer-form-comparison} groups queries by answer form. Long-form queries (daily briefings, project status reports) require synthesizing information across many documents into a coherent narrative. Short-form queries (meeting recaps, action item rollups, collaborator lookups, etc.) target specific facts or events.

\begin{table}[h]
\centering
\small
\setlength{\tabcolsep}{4pt}
\begin{tabular}{l l c c c c c c c c}
\toprule
\textbf{Answer Form} & \textbf{System} & \textbf{Comp.} & \textbf{Sound.} & \textbf{Relev.} & \textbf{Read.} & \textbf{Faith.} & \textbf{Retr.} & \textbf{Final} & \textbf{Overall} \\
\midrule

\textbf{Long Form} & ReAct & 0.496 & 0.762 & 0.757 & 0.894 & 0.909 & 0.551 & 0.717 & 2.724 \\
 & Onyx & 0.428 & 0.823 & 0.619 & 0.810 & 0.866 & 0.406 & 0.679 & 2.667 \\

\textbf{Short Form} & ReAct & 0.665 & 0.826 & 0.571 & 0.949 & 0.917 & 0.384 & 0.765 & 2.993 \\
 & Onyx & 0.601 & 0.876 & 0.680 & 0.831 & 0.920 & 0.381 & 0.766 & 3.220 \\

\bottomrule
\end{tabular}
\vspace{1.5mm}
\caption{Comparison across query types}
\label{tab:answer-form-comparison}
\vspace{-3mm}
\end{table}

The answer breakdown reveals distinct failure modes across query structures. 
Long-form queries (daily briefings and project status reports) require synthesizing information distributed across many documents into a coherent narrative. Both systems perform substantially worse on these queries than on short-form queries, with holistic scores below 2.8 and noticeably reduced completeness. ReAct achieves a higher final score on long-form queries (0.717 vs.\ 0.679), primarily due to stronger completeness (0.496 vs.\ 0.428) and retrieval accuracy (0.551 vs.\ 0.406). These results suggest that both systems struggle to reliably aggregate all relevant context across long-horizon enterprise workflows.

On short-form queries, final scores converge almost exactly (0.765 vs.\ 0.766), but the metric breakdown reveals a consistent completeness against relevance tradeoff. ReAct achieves higher completeness (0.665 vs.\ 0.601), while Onyx attains substantially higher relevance (0.680 vs.\ 0.571). This suggests that ReAct tends to include broader contextual information, whereas Onyx produces more targeted and concise responses.  A detailed breakdown across individual query types is provided in Table~\ref{tab:query-type-full-results}.

\section{Conclusion and limitations}
\label{sec:conclusion}
In this work, we introduced \textsc{WinSyn}, an automated pipeline for constructing realistic, diverse synthetic enterprise email corpora designed to evaluate question–answering systems. \textsc{WinSyn}'s design is motivated by the rich structure of enterprise data. By grounding all generated artifacts, ranging from hierarchical project structures to email communications in concise daily activity logs, ensures that gold answers remain verifiable and auditable. The pipeline’s staged design, combined with reflection–refinement loops and programmatic validators, enables the generation of coherent, temporally consistent data with rich complexity. 
Empirical evaluation on multiple datasets demonstrates that even strong agentic baselines struggle to achieve high performance, highlighting persistent challenges in retrieval, reasoning, and grounding under realistic enterprise conditions. We hope this work provides both a practical framework for scalable dataset generation and a rigorous testbed for advancing enterprise search, retrieval and deep research  systems.

The current scope of our dataset is limited to emails. While emails are already capable of representing a large part of communication and thus capture most of the complexity, in real-world workplace settings, communication and work artifacts are distributed across multiple channels, including meetings, chats, code repositories, ticketing systems, and other collaborative tools. Regarding realism, we have not quantified or measured the distributional similarity of the dataset to actual enterprise data (what we termed statistical realism). Nor did we deeply explore other aspects of realism mentioned earlier. We did not include external interactions (e.g., customer facing interactions), and did not work with company-specific culture or jargon. More types of queries can be easily incorporated in our pipeline but all our queries are answerable; introducing unanswerable queries \cite{choubey2025herb} is also a useful diagnostic tool. While we have thought about our dataset as a benchmark used for evaluation, it could also be used for finetuning, perhaps with some enhancements. Freedom in the choice of the seed document can help here. Moreover, practitioners could specify specific failure modes encountered in practice, and the synthetic data can incorporate them. In a different type of limitation, our evaluations did not make use of the sentence-level attribution that our pipeline constructs. This requires the baseline agents to construct such attribution which can then be used to check grounding of the answers generated by the baseline agents. 

Addressing these limitations and scaling up the dataset in terms of size (bigger organization with more complex dynamics) and time period are left to future work. 
We believe that our approach provides a solid foundation for this.

\newpage
\bibliographystyle{plainnat}
\bibliography{neurips_2026}

\newpage
% \input{checklist}

% References follow the acknowledgments in the camera-ready paper. Use unnumbered first-level heading for
% the references. Any choice of citation style is acceptable as long as you are
% consistent. It is permissible to reduce the font size to \verb+small+ (9 point)
% when listing the references.
% Note that the Reference section does not count towards the page limit.
% \medskip

% {
% \small

% [1] Alexander, J.A.\ \& Mozer, M.C.\ (1995) Template-based algorithms for
% connectionist rule extraction. In G.\ Tesauro, D.S.\ Touretzky and T.K.\ Leen
% (eds.), {\it Advances in Neural Information Processing Systems 7},
% pp.\ 609--616. Cambridge, MA: MIT Press.

% [2] Bower, J.M.\ \& Beeman, D.\ (1995) {\it The Book of GENESIS: Exploring
%   Realistic Neural Models with the GEneral NEural SImulation System.}  New York:
% TELOS/Springer--Verlag.

% [3] Hasselmo, M.E., Schnell, E.\ \& Barkai, E.\ (1995) Dynamics of learning and
% recall at excitatory recurrent synapses and cholinergic modulation in rat
% hippocampal region CA3. {\it Journal of Neuroscience} {\bf 15}(7):5249-5262.
% }

%%%%%%%%%%%%%%%%%%%%%%%%%%%%%%%%%%%%%%%%%%%%%%%%%%%%%%%%%%%%
\appendix
% \localtableofcontents

\section*{Appendix}
% \addcontentsline{toc}{section}{Appendix Index}

% \begin{itemize}[leftmargin=*]
% \item \hyperref[app:B]{Methodology}
% \item \hyperref[app:C]{Dataset Code Examples}
% \end{itemize}

\section{Pipeline details}
\label{app:B}
In this section we will outline our methodology for dataset creation. This section is divided into sub-sections each of which tackles a different part of the dataset generation pipeline.

We propose a hierarchical, multi-stage pipeline for generating realistic synthetic enterprise communication datasets based on the initial real-world or synthetic seed. The pipeline transforms a single unstructured seed document into a richly interconnected corpus of workplace artifacts, project, epics, task dependency graphs, employee diaries, email threads and question and answer pairs while maintaining end-to-end factual traceability. Each generation stage is paired with structured reflection, validation, and iterative refinement loops that leverage a dedicated validation model to enforce coverage, consistency, and groundedness constraints before proceeding to downstream stages.

The pipeline architecture follows a top-down decomposition strategy where a macro-level project structure is established first, then progressively refined and instantiated into fine-grained daily artifacts such as emails. Concretely, a single multi-stage process converts the seeding information into a workplace interaction corpus each conditioned on prior outputs and validated by programmatic and LLM-based checks. This design mirrors real organizational workflows and ensures that all generated content remains coherent across temporal, organizational, and informational dimensions.

\subsection{Overview and Notation}
\label{sec:method:overview}

Let $N$ denote an input seed document (which can be a technical report, a change/update log, blog post, newsletter, etc.). 
From this a project time window $[t_0, t_1]$ is extracted. 

The pipeline produces, in order, a synthetic company context $C$ and
an employee set $U$ with a reporting hierarchy graph $H$, an epic DAG
$\mathcal{G}_E = (V_E, A_E)$, a per-epic
task DAG $\mathcal{G}_T^{(e)} = (V_T^{(e)}, A_T^{(e)})$ for each
$e \in V_E$ , a daily diary $D_{\tau,d}$ for
every task $\tau$ and business day $d$,
a QA set $\mathcal{Q}$, communication
artifacts which are emails conditioned on $\mathcal{Q}$, and a sentence-level attribution map
$\alpha : \mathcal{Q} \to 2^{\mathcal{S}}$ where $\mathcal{S}$ is the
set of indexed sentences extracted from diaries and communications. \navin{The point of this attribution map is not clear.} All structured outputs are emitted against fixed JSON schemas; reflection and refinement prompts are
versioned alongside the schemas they govern, and all LLM calls share
a single system prompt to preserve stylistic consistency across
artifacts.

We follow the broad three design patterns across the pipeline at different stages. 

\begin{itemize}
\item \textbf{Grounding by construction.} Every artifact such as \textit{epic}, \textit{task} and even \textit{emails} is grounded in the artifacts of the predecessor stages with reflection and refinement loops described below. \navin{In the main paper we are briefly defining the terminology of epics and tasks. Can include that here too.}
    
\item \textbf{Validate Reflect and Refine loop.} Each stage is followed by programmatic validators and LLM based reflection and refinement steps until the output is satisfactory until a preset number of attempts is exhausted. 
    
\item \textbf{Conditioned Generation.} Gold answers are frozen \emph{before} any email communication is generated. Emails only add details that do not change the gold answers.
This turns the consistency problem between answers and supporting evidence into a tractable conditional generation problem.
\end{itemize}

Below we expand on the pipeline stages and discuss the design choices made therein.

\subsection{Pipeline Stages}
\label{sec:method:stages}

\subsubsection{Company, Employees and Projects}
\label{sec:method:roster}

We start with a real or fictional company and its metadata (industry,
size, domain) along with a seed document describing a project $N$, to anchor the 
downstream artifacts in a coherent manner. We then generate a roster
$\mathcal{U}$ of $20$--$25$ employees together with a separate plain-text
reporting hierarchy. Each employee $u \in \mathcal{U}$ carries a name,
role, team, responsibilities, background, personality attributes,
focus areas and free-text skill descriptors. The skill descriptors and
the hierarchy are later used as soft constraints during epic and task
assignment, and the hierarchy additionally identifies the senior-most
personnel that anchor the opening epic.

\subsubsection{Hierarchical Epic Generation}
\label{sec:method:epics}

\ananya{should we define epic?}

\ananya{add paragraph about future reference label in epic segment}

\navin{(I will take care of this comment.) This is done in the main paper and if we want to redo it in the appendix it needs to be done earlier in the appendix, not here. We also need to include some rough numbers like how many epics.}
% We follow the terminology of \textit{agile methodology of software development} although our technqiue is general and can be readily applied in other situations as well. We borrow the terms \textit{epics} and \textit{tasks} from there
% We represent the epic structure as a directed acyclic graph
% $\mathcal{G}_E = (V_E, \mathcal{R}_E)$, where $V_E$ denotes the set of
% epics and $\mathcal{R}_E$ encodes precedence relations. 

We follow the terminology of agile software development, although our framework is general and applicable beyond software engineering workflows. In particular, we use the notions of \emph{epics} and \emph{tasks}. An \emph{epic} represents a high-level product objective or feature area, while \emph{tasks} correspond to the concrete implementation activities required to realize that objective.

We represent the epic structure as a directed acyclic graph (DAG) $G_E = (V_E, R_E)$, where each node $e \in V_E$ denotes an epic and each directed edge $(e_i, e_j) \in R_E$ encodes a precedence relation indicating that epic $e_i$ must precede epic $e_j$. Each epic additionally contains fields such as a short textual \texttt{description} summarizing the feature objective and a priority or complexity estimate used for planning and scheduling. The \texttt{dependency\_notes} capture coarse-grained project structure and dependencies on the predecessor epics. Each epic also contains \texttt{start\_date}, \texttt{end\_date} and \texttt{assigned\_employees} which help with further downstream scoping.

Each epic $e \in V_E$ additionally contains a set of associated tasks $T_e = \{t_1, t_2, \dots, t_n\}$, which define the finer-grained implementation steps within the epic. While real-world planning often involves richer constraints, in this work we focus on precedence dependencies represented by $R_E$.

% Generating the full graph $$G_E$$ in a single LLM call often produces an undifferentiated project structure. We therefore construct it incrementally in three sub-stages.

\navin{(I will take care of this comment.) What this precedence relation means needs elaboration. Actually, we are working with a simple version of this and there are more sophisticated constraints too that arise in practice but we won't discuss those.} Generating
$\mathcal{G}_E$ in a single LLM call collapses the project
into one undifferentiated phase; we therefore construct it in three
sub-stages. 

Sub-stage~1 emits a single \emph{opening} epic $e_1$, explicitly
constrained to be a kickoff and planning phase and to include the
organization's senior employees. This anchors governance and
provides a unique entry node.

Sub-stage~2 emits a configurable set of \emph{intermediate} execution
epics conditioned on $e_1$, with explicit instructions not to
regenerate the opening epic or pre-empt closure.

Sub-stage~3 emits a single \emph{closing} epic conditioned on all
preceding epics, scoped to delivery, communication, and ownership
transition rather than mere hand-off.

\subsubsection{Task Decomposition}
\label{sec:method:tasks}

Each $e \in V_E$ is decomposed into a task-level directed acyclic graph
$\mathcal{G}_T^{(e)} = (V_T^{(e)}, \mathcal{R}_T^{(e)})$. Each task $\tau \in V_T^{(e)}$ specifies a title, a scoped
description, assigned employees drawn from $U(e)$ (the set of employees
associated with epic $e$), a business-day timeline, intra- and
inter-epic dependencies, and explicit \texttt{expected\_outputs} and
\texttt{unlocks} fields. \navin{Explain what these fields are. Maybe give an example too.} Concretely, \texttt{expected\_outputs} enumerates the named artefacts a task must emit (e.g., \texttt{expected\_outputs:\ ["risk\_register.csv",\ "mitigation\_plan.md"]}), while \texttt{unlocks} lists the downstream task ids that become eligible to run once those artefacts exist (e.g., \texttt{unlocks:\ ["mitigation\_planning",\ "exec\_review"]}). Together they act as a \textit{contract}: a scheduler verifies hand-offs with a deterministic set-membership check that is every successor's prerequisites must appear in some predecessor's \texttt{expected\_outputs} thus catching dangling references and orphan deliverables without invoking an LLM judge.
\navin{more elaboration needed in the previous paragraph. The kind of information we generate for tasks has a counterpart for epics but it's missing in the epics section.}

The first task of every epic is constrained to be a kickoff and the last to be a closure or, when $e$ is the closing epic, the project
closure so that every epic has a well-defined entry and exit
deliverable. \navin{Just like the last task of the last epic is special, isn't the first task of the first epic special?}

As an optional alternative to LLM assigned dates, a deterministic
forward scheduler computes start and end dates by topological traversal
of $\mathcal{G}_T^{(e)}$ under a per-employee concurrency cap, then
anchors the resulting schedule to $t_1$ via a uniform shift; this is
useful when strict temporal feasibility is required.

\subsubsection{Daily Diary Synthesis}
\label{sec:method:diaries}

Diaries constitute the canonical record of \emph{what happened} on each
business day and form the substrate against which downstream QA is
grounded. Let $\mathcal{D}$ denote the set of business days within the
time window $[t_0, t_1]$, and let
$\mathrm{timeline} : V_T^{(e)} \rightarrow 2^{\mathcal{D}}$ map each
task to its scheduled execution days (a time interval).

For each task $\tau \in V_T^{(e)}$ and each day
$d \in \mathrm{timeline}(\tau)$, we emit a structured diary entry
$D_{\tau,d}$ comprising fields for work summary, per-employee detailed
activities, progress notes, blockers (with reporter), resolutions
(with resolver), collaboration interactions (with participant lists).
\navin{An example would help a lot here.}

Generation is conditioned on the position of $\tau$ within
$\mathcal{G}_T^{(e)}$: predecessor deliverables, successor
requirements, and cross-task hand-off notes are injected into the
prompt context. This conditioning enables downstream artifacts such as
emails to refer to prior days without contradiction.

Per-task diary entries are then aggregated into
per-day, per-epic views via $D_{e,d} = \bigcup_{\tau \in V_T^{(e)}} D_{\tau,d}$ which are used for downstream consumption.

\subsubsection{Question and Answer Generation}
\label{sec:qna-generation}

We generate a set of question--answer pairs $Q$ before any communication artifact, so that gold
answers are grounded solely in the diary corpus. This ordering is central: it allows answers to act
as constraints on subsequent generation.

We support two complementary generation flows. The first produces questions tied
to an employee's recent work: a \textbf{daily briefing} scoped to the epics active for that
employee, and a \textbf{project status} overview scoped to the full project asked from a leadership's perspective We have two fixed question templates for this type ("What should I work on today?", "Generate a report about the project"). The
second generation flow produces questions that require identifying evidence across multiple
tasks, or epics, for example, tracing ownership, summarising
collaborative work within a time window, comparing progress across epics, or preparing a meeting
briefing, identifying open work, blockers, tracking handoffs. More details on the question types are added in Table \ref{tab:question-types}. Here the question types are fixed but the individual questions are generated from the LLM.

Both flows share the same downstream quality gate but differ in how context is assembled and how
answers are produced.

\paragraph{Report type questions (daily briefing and project status).}
Let $D$ denote the set of business days and $U$ the set of employees. A sampling module selects
pairs $(u, d) \in U \times D$ such that $u$ is assigned to at least one active epic on day $d$ and
there exist sufficient diary entries for $u$ before $d$.
Sampling is deterministically seeded for reproducibility.

For each sampled pair, the model receives a \emph{project snapshot} (a structured summary of every
active epic's goals, timeline, and current task status) together with the employee's persona and a
window of diary entries. For daily briefing questions the window covers the employee's own active
tasks; for project status questions it covers all active tasks across the project. In both cases, recent
entries (most recent 3 days for daily briefing, 15 days for project status) are provided in full detail,
while earlier entries are summarised to not make the model overburdened with lot of context.

Answers are produced in two passes. In Pass~1 the model generates an initial answer from this context, citing the diary entries it relies on. In Pass~2 every cited entry is expanded to
full detail and the answer is regenerated, ensuring that the final response is grounded in complete
rather than summarised evidence.

Each candidate answer is then evaluated by an LLM-based grounding critic that identifies
ungrounded claims. An answer is accepted if and only if it is non-empty, fully grounded, and covers
the relevant diary evidence. Answers containing any ungrounded claims are rejected; the critic's
feedback, listing each unsupported claim with a reason and suggested action, is incorporated into
the prompt and generation is retried up to a fixed budget. If no valid answer is produced within this
budget, the instance is discarded with a logged warning.

\paragraph{Short form questions.}
For these queries, context is assembled by traversing the task dependency graph to extract
chains, sequences of tasks linked by dependency edges, potentially spanning multiple epics. The
pipeline enumerates candidate chains and preferentially retains longer ones, since these tend to
involve more employees, richer handoff structure, and a wider spread of diary evidence. Each chain
$c$ is scored by length of the dependency chain and total day span, and the top-$k$ chains are retained. 

For question-answer generation cycles through selected chains, the model receives the
question-type spec (details about what type of question to generate), the chain context (task and epic details) for structural understanding, the diary entries before a sampled \texttt{asked\_on} date, and the project snapshot. It is instructed to ground all
factual claims exclusively in the diary entries and to phrase questions and answers in the register of
workplace communication.

Each candidate is then evaluated by an LLM-based verifier that expands the evidence set
by identifying related tasks not in the original context, then checks diary-grounded entailment. The
verifier produces a verdict $v \in \{\textsc{verified},\;\textsc{ambiguous},\;\textsc{not\_verified}\}$
and a recommendation $r \in \{\textsc{keep},\;\textsc{rewrite\_answer},\;\textsc{reject}\}$.
Verified candidates are accepted; candidates with fixable issues are minimally rewritten; all others
are rejected.

\begin{table}[ht]
\centering
\small
\begin{tabular}{p{3.8cm} p{4cm} p{4cm} p{0.8cm}}
\toprule
\textbf{Question Type} & \textbf{Description} & \textbf{Example Question} & \textbf{Answer Length (Sentences)} \\
\midrule
communication\_traceback & Reconstruct the context behind a specific ask, promise, or email thread tied to project work & What did I promise the support team in my last handoff email? & 2--3 \\

multi\_hop\_people\_queries & Attribute ownership, responsibility, or expertise to individuals across work items and handoffs & Who owns the dashboard rollout? / Who is driving the API migration? & 2--4 \\

action\_item\_rollup & Enumerate open follow-ups, unresolved tasks, and downstream asks from recent project activity & What follow-up tasks are still open on this rollout? & 3--5 \\

artifact\_handoff\_locator & Locate artifacts, packages, or runbooks produced or handed off, and identify who owns or consumes them & What artifacts were handed off from this workstream and who are they for? & 3--5 \\

collaborator\_network\_analysis & Map who collaborated with a given person on a topic or workstream across diary entries and handoffs & Who worked with the main deployment lead on this effort? & 5--6  \\

decision\_and\_feedback\_recap & Recover feedback, decisions, and follow-ups from a review, handoff, or discussion thread & What feedback and decisions came out of the architecture review? & 5--6 \\

meeting\_prep & Summarize unresolved work, progress, and blockers to prepare for an upcoming sync, review, or 1:1 & What should I review before the planning discussion for this rollout? & 5--6 \\

time\_and\_collaboration\_insights & Surface work, blockers, or collaboration patterns within a specific time window or milestone & What did I work on last week? / Who do I interact with most on this workstream? &  \\

multi\_domain\_catch\_up & Synthesize status, risks, and tasks across a project or dependency-heavy workstream & What's the current status of this rollout? & 7--8 \\

project\_reflection & Retrospective on accomplishments, effort themes, and progress across a project window & What did I accomplish in this workstream last month? & 7--8 \\

daily\_briefing & Predictive daily focus briefing grounded in active epics and recent diary history & What should I focus on at work today? & 4 sections, $\sim$13  \\

project\_status & Holistic project health report covering all workstreams, risks, blockers, and upcoming milestones & What is the current status of the complete project? & 5 sections, $\sim$13 sentences + table \\
\bottomrule
\end{tabular}
\vspace{2mm}
\caption{Question types used in the dataset, ordered by increasing answer complexity and length.}
\label{tab:question-types}
\end{table}

\subsubsection{Communication Synthesis with Locked QA}
\label{sec:method:entities}

\navin{Need to discuss this section too.}
Communication artifacts are synthesised last, conditioned on the task
graphs $\mathcal{G}_T^{(e)}$, the diary corpus $\{D_{\tau,d}\}$, and a
\emph{locked} view of the question and answer set $\mathcal{Q}$. The lock
ensures that all generated communications remain consistent with the
previously accepted answers, which act as fixed constraints.

For each task $\tau \in V_T^{(e)}$ and day $d \in \mathrm{timeline}(\tau)$,
we construct a conditioning subset
\[
\mathcal{Q}_{\tau,d} = \{\, q \in \mathcal{Q} \mid d_q > d,\;
e(q) = e,\; u(q) \in U(\tau) \,\},
\]
where $d_q$ and $u(q)$ denote the date and employee associated with
question $q$, $e(q)$ denotes the referenced epic, and $U(\tau) \subseteq
\mathcal{U}$ is the set of employees assigned to task $\tau$. Intuitively,
$\mathcal{Q}_{\tau,d}$ contains only those questions whose answers must
remain supportable by communications generated at $(\tau,d)$, without
leaking future information.

Because emails are required to be consistent with
answers they will eventually \emph{support} rather than the other way
around, this lock-context mechanism prevents drift between evidence and
answers without requiring post hoc reconciliation.

For each task $\tau$, we generate exactly one email thread
$\mathcal{L}^{(\tau)}$, processed chronologically over the days in
$\mathrm{timeline}(\tau)$. Days with no blockers, no collaboration, and
no progress notes in $D_{\tau,d}$ are skipped to avoid content-free
messages.
\ananya{should we remove meeting, chat?}
\amey{Removed the meeting content}

Each artifact additionally stores a pointer
$\mathrm{ref}(\cdot) \rightarrow D_{\tau,d}$ to its source diary entry,
providing the structural backbone for downstream attribution.
\navin{An important part here was to }

\ananya{looks like it is repeated here}

\subsubsection{Sentence Level Attribution}
\label{sec:method:attribution}

\navin{Like in several other places, this section launches into technical details without saying first where it's going. First say what the destination is and then show the path.}
Once all artifacts are committed, we compute a sentence level
attribution map $\alpha$. \navin{I didn't quite follow this splitting.} Let $\mathcal{S}$ denote the set of all
sentences obtained by decomposing diary and communication artifacts.
Sentences are produced via a three-pass splitter: (i) segmentation
along structural boundaries, (ii) splitting of long segments using
sentence terminators, and (iii) merging of short fragments to preserve
semantic coherence. Each sentence $s \in \mathcal{S}$ is indexed with
metadata
$(\texttt{epic\_id}, \texttt{task\_id}, \texttt{date},
\texttt{section}, \texttt{parent\_artifact\_id})$.

Let $\mathcal{S}_A \subseteq \mathcal{S}$ denote answer sentences,
$\mathcal{S}_D$ diary sentences, and $\mathcal{S}_C$ communication
sentences. Attribution is computed in two LLM-mediated passes composed
transitively through the diary layer:
\[
\alpha = \alpha_{A \rightarrow D} \circ \alpha_{D \rightarrow C}.
\]

The first pass $\alpha_{A \rightarrow D} \subseteq
\mathcal{S}_A \times \mathcal{S}_D$ maps answer sentences to
\emph{substantive evidence} diary sentences, where substantive evidence
is defined as sentences expressing concrete actions, decisions, or
outcomes (excluding generic or preparatory statements). The second pass
$\alpha_{D \rightarrow C} \subseteq \mathcal{S}_D \times \mathcal{S}_C$
maps diary sentences to supporting communication sentences, with
non-informative elements such as email subject lines and greetings
filtered out.

Routing attribution through the diary layer is deliberate: direct
mapping from $\mathcal{S}_A$ to the full communication corpus
$\mathcal{S}_C$ would be combinatorially expensive and would discard
the provenance chain required for downstream evaluation.

% \subsection{Reflection-Refinement Loops}
% \label{sec:method:refinement}

% Each stage of the pipeline that emits structured artifacts (specified JSON schema) is followed by one or more
% \emph{reflection--refinement} loops of the form
% \begin{equation*}
% A^{(0)} \xrightarrow{\text{reflect}} R^{(0)}
% \xrightarrow{\text{validate}} v^{(0)}
% \xrightarrow{\text{refine if } v^{(0)} = \bot} A^{(1)} \;\cdots,
% \end{equation*}
% where $A^{(i)} \in \mathcal{A}$ denotes the artifact instance at
% iteration $i$, $R^{(i)}$ is an LLM-generated critique evaluated against
% an explicit rubric, and $v^{(i)} \in \{\top, \bot\}$ is the verdict of a
% deterministic validator.

% We deliberately separate the critic from the generator (via distinct
% prompt templates and, where possible, separate model invocations) to
% reduce the risk of self-reinforcing errors in which a model endorses
% its own outputs. The refinement operator produces
% $A^{(i+1)} = \mathrm{refine}(A^{(i)}, R^{(i)})$ when
% $v^{(i)} = \bot$.

% A loop terminates when either $v^{(i)} = \top$ or a per-stage iteration
% cap $I_s$ is reached. Intermediate states $\{A^{(i)}\}$ are checkpointed
% to enable recovery from transient failures, and resume points are
% exposed at each loop boundary (e.g., post-coverage validation,
% dependency validation, or post-refinement).

% The combination of a soft critic and a hard validator ensures that the
% loop is finite in practice: execution halts immediately once the
% validator accepts, even if the critic continues to suggest further
% improvements.

% \ananya{looks like it is repeated here}

\subsubsection{Distractor Generation}
\label{sec:distractors}

After all question and answer pairs have been finalised, we inject \emph{distractor} artifacts into the emails that are plausible but irrelevant or misleading with respect to the gold answers. Because distractors are generated post the QA generation, gold answer semantics are preserved by construction, enabling controlled evaluation of model grounding and faithfulness.
\paragraph{Distractor types.}
We generate three broad categories of distractors:

\begin{enumerate}
  \item \emph{Process noise: } Routine workplace communications unrelated to project work (out-of-office replies, HR announcements, IT maintenance notices, social events). These are safe by design: they contain no factual claims about project outcomes.
  \item \emph{Correction chains: } Seeded from the existing corpus, these threads introduce an initial misstatement that is subsequently corrected, with the final state matching ground truth. They test whether a model can reconcile conflicting information within a thread.
  \item \emph{Tangential-project communications: } Messages from separate teams using overlapping vocabulary (e.g., ``Dashboard 2.0'' vs.\ ``Reporting Dashboard''). These test whether a model respects project scope boundaries when terminology is shared.
\end{enumerate}

\paragraph{Generation and validation.}
For each QnA target, the pipeline extracts relevant persons, a temporal lookback window (2--7 days before \texttt{asked\_on}), and key facts. Correction chains are seeded from a real artifact within this window; the LLM then generates a mistake and correction pair whose resolution aligns with the gold answer. Tangential-project distractors are generated contextually with a fabricated but plausible parallel team. Each distractor is validated to ensure no gold answer would change in its presence: process noise is auto-approved, while correction chains and tangential-project distractors undergo LLM-based comparison against the original answer. Accepted distractors are tagged and merged into the corpus.

\subsection{Reflection-Refinement Loops}
\label{sec:method:refinement}

Each stage that emits structured artifacts is followed by one or more
\emph{reflection--refinement} loops of the form
\begin{equation*}
A^{(0)} \xrightarrow{\text{reflect}} R^{(0)}
\xrightarrow{\text{validate}} v^{(0)}
\xrightarrow{\text{refine if } v^{(0)} = \bot} A^{(1)} \;\cdots,
\end{equation*}
where $A^{(i)} \in \mathcal{A}$ denotes the artifact instance at
iteration $i$, $R^{(i)}$ is an LLM-generated critique evaluated against
an explicit rubric, and $v^{(i)} \in \{\top, \bot\}$ is the verdict of a
deterministic validator.

We deliberately separate the critic from the generator (via distinct
prompt templates and, where possible, separate model invocations) to
reduce the risk of self-reinforcing errors in which a model endorses
its own outputs. The refinement operator produces
$A^{(i+1)} = \mathrm{refine}(A^{(i)}, R^{(i)})$ when
$v^{(i)} = \bot$.

A loop terminates when either $v^{(i)} = \top$ or a per-stage iteration
cap $I_s$ is reached. Intermediate states $\{A^{(i)}\}$ are checkpointed
to enable recovery from transient failures, and resume points are
exposed at each loop boundary (e.g., post-coverage validation,
dependency validation, or post-refinement).

% \paragraph{Generation and validation.}
% For each QnA target, the pipeline extracts relevant persons, a temporal lookback window (2--7 days before \texttt{asked\_on}), and key facts. Correction chains are seeded from a real artifact within this window; the LLM then generates a mistake and correction pair whose resolution aligns with the gold answer. Tangential-project distractors are generated contextually with a fabricated but plausible parallel team. Each distractor is validated to ensure no gold answer would change in its presence: process noise is auto-approved, while correction chains and tangential-project distractors undergo LLM-based comparison against the original answer. Accepted distractors are tagged with \texttt{is\_distractor\,=\,True} and merged into the corpus.
\subsubsection{Validators and Rubrics}
\label{sec:method:validators}

At the epic level, we run four independent reflection--refinement loops
with explicit rubrics. The \emph{seed document} rubric enforces
that every factual span $s \in \mathcal{S}_N$ (where $\mathcal{S}_N$
denotes the set of textual spans extracted from $N$) is associated with
at least one epic $e \in V_E$, and that no epic introduces claims not
grounded in $N$. The \emph{future-reference labeling} rubric ensures
that segments tagged \texttt{future\_reference} correspond to content
that lies strictly outside the epic time window $[t_0^e, t_1^e]$, where these are the start and end times of the epic. The
\emph{dependency} rubric enforces acyclicity of $\mathcal{G}_E$,
predecessor consistency, and timeline feasibility; if the critical path
exceeds the available window, the horizon $[t_0, t_1]$ may be extended
by a configurable margin and the loop re-run. The \emph{employee
assignment} rubric enforces load balance across $\mathcal{U}$ and
consistency with the reporting hierarchy.

After refinement, an \emph{anchor validation} pass enforces structural
invariants: the first epic $e_1$ is a planning epic with no predecessors,
the final epic $e_n$ is a closure epic, and identifiers are sequential.
Violations at this stage are treated as hard errors, as they break
assumptions required by downstream stages.

At the task level, we run analogous loops for each epic $e \in V_E$ in
topological order over $\mathcal{G}_E$. The \emph{coverage} rubric
ensures that task-level content collectively accounts for the epic's
assigned seed document spans. The \emph{hand-off} rubric enforces that for
each task $\tau \in V_T^{(e)}$, the declared
\texttt{expected\_outputs} satisfy the dependency requirements of its
successors in $\mathcal{G}_T^{(e)}$, and that the final task closes the
epic (or the entire project if $e = e_n$). The \emph{dependency} rubric
enforces acyclicity of $\mathcal{G}_T^{(e)}$, and the \emph{assignment}
rubric ensures consistency between task-level assignments
$U(\tau) \subseteq \mathcal{U}$ and the epic-level allocation $U(e)$.

A final task anchor-repair loop performs up to three LLM-assisted
refinement attempts with exponential backoff before failing.

\subsection{Validation and Quality Assurance Summary}
\label{sec:method:qa}

The validation and quality assurance pipeline operates as a layered set of checks spanning artifacts, dependencies, and generated outputs. At a high level, the framework combines programmatic verification (e.g., fuzzy matching, structural checks, and timeline arithmetic) with iterative LLM-based reflection to ensure both syntactic and semantic consistency. These validations are applied at multiple stages of the pipeline, from initial seed document coverage through epic and task graph refinement, to final answer generation and attribution.

Crucially, the system enforces both coverage and grounding guarantees. Coverage-oriented checks ensure that all required artifacts (e.g. epics, tasks, and references) are adequately represented and aligned, while grounding mechanisms such as claim-level verification and sentence-level attribution provide traceability back to source documents. Together, these complementary validation layers enable robust error detection and correction, ensuring that the final outputs remain consistent, complete, and verifiably grounded.

\begin{table}[h]
\centering
\small
\renewcommand{\arraystretch}{1.15}
\begin{tabularx}{\textwidth}{
@{}
>{\raggedright\arraybackslash}p{0.22\textwidth}
>{\raggedright\arraybackslash}p{0.20\textwidth}
>{\raggedright\arraybackslash}X
>{\raggedright\arraybackslash}p{0.22\textwidth}
@{}}
\toprule
\textbf{Validation} & \textbf{Scope} & \textbf{Method} & \textbf{Trigger} \\
\midrule
seed document coverage & $V_E \leftrightarrow \mathcal{S}_N$ & Programmatic fuzzy + semantic match & Each iteration; final epics \\
Future-reference audit & $\mathcal{S}_N$ & LLM reflection + programmatic patch & After coverage refinement \\
Epic dependency feasibility & $\mathcal{G}_E$ & LLM reflection + timeline arithmetic & During dependency refinement \\
Epic anchor invariants & $V_E$ & LLM reflection (hard error) & After epic refinement \\
Task coverage & $V_T^{(e)} \leftrightarrow V_E$ & Programmatic fuzzy match & After task graphs final \\
Task anchor / hand-off & $\mathcal{G}_T^{(e)}$ & LLM reflection + repair loop & After task refinement \\
QA grounding & $\mathcal{Q}$ & LLM claim-level fact check (retry loop) & After each answer \\
Sentence-level attribution & $\alpha_{A \rightarrow D} \circ \alpha_{D \rightarrow C}$ & Two-step LLM mapping via diary join & After all generation \\
\bottomrule
\end{tabularx}
\vspace{2mm}
\caption{Validation mechanisms layered across the pipeline.}
\end{table}

\section{Additional Experiments and Results}

To supplement the results in the Section \ref{sec:references}, we provide a more detailed view of the \textit{per-query metrics} beyond the aggregated results provided in the main text.

\textbf{Retriever} For the ReAct agent a hybrid retriever is used. The retriever is a BM25 + dense vector ensemble implemented using LlamaIndex, retrieving up to $k = 60$ document chunks per query. Documents are split into chunks of up to 256 tokens using a sentence-aware splitter (\texttt{LlamaIndex SentenceSplitter}), with a 32-token overlap between consecutive chunks to preserve local context across boundaries. The agent iterates for up to 10 reasoning steps, invoking the retriever and auxiliary metadata tools (e.g., employee lookup, role search) as needed. The Onyx Retrieval mirrors Onyx's proprietary Vespa hybrid profile. The email bodies chunked at 1{,}600 characters (200-character overlap), $\alpha{=}0.75$, $\mathrm{tcr}{=}0.3$, $\lambda{=}0.05$, and top-$k{=}50$. Relative to the original Onyx agent, we swap the Vespa search with an equivalent search system over the on-disk \texttt{.eml} sandbox.

The Table~\ref{tab:query-type-full-results} provides a fine-grained breakdown of performance across individual query types. In addition to the answer-quality metrics discussed in the main paper, the table also includes retrieval-oriented metrics such as MRR, Recall, and HR@25 to provide additional visibility into retrieval behavior across different query structures. We group query types into long-form and short-form categories to highlight differences between synthesis-heavy tasks (e.g., daily briefings and project status reports) and more localized factual retrieval tasks (e.g., collaborator lookup, action item retrieval, and meeting recap queries).

\begin{table*}[t]
\centering
\scriptsize
\setlength{\tabcolsep}{4pt}
\begin{tabular}{p{3.2cm} l c c c c c c c c c c c}
\toprule
\textbf{Query Type} & \textbf{System} & \textbf{Comp.} & \textbf{Sound.} & \textbf{Relev.} & \textbf{Read.} & \textbf{Faith.} & \textbf{Retr.} & \textbf{Overall} & \textbf{Final} & \textbf{MRR} & \textbf{Recall} & \textbf{HR@25} \\
\midrule

\multicolumn{13}{l}{\textbf{Long Form}} \\

Daily Briefing & ReAct & 0.477 & 0.712 & 0.792 & 0.867 & 0.908 & 0.540 & 2.803 & 0.698 & 0.480 & 0.833 & 0.955 \\
 & Onyx & 0.297 & 0.712 & 0.725 & 0.873 & 0.820 & 0.323 & 2.409 & 0.622 & 0.172 & 0.574 & 0.818 \\

Project Status & ReAct & 0.519 & 0.820 & 0.716 & 0.926 & 0.909 & 0.563 & 2.632 & 0.738 & 0.389 & 0.758 & 1.000 \\
 & Onyx & 0.174 & 0.784 & 0.798 & 0.947 & 0.915 & 0.412 & 2.316 & 0.639 & 0.355 & 0.508 & 0.947 \\

\midrule
\multicolumn{13}{l}{\textbf{Short Form}} \\

Action Item Rollup & ReAct & 0.470 & 0.709 & 0.283 & 1.000 & 0.832 & 0.248 & 1.750 & 0.621 & 0.223 & 1.000 & 1.000 \\
 & Onyx & 0.135 & 0.575 & 0.660 & 0.900 & 0.855 & 0.270 & 1.750 & 0.530 & 0.163 & 0.923 & 1.000 \\

Artifact Handoff Locator & ReAct & 0.276 & 0.572 & 0.457 & 0.800 & 0.942 & 0.213 & 2.500 & 0.544 & 0.142 & 1.000 & 1.000 \\
 & Onyx & 0.687 & 0.857 & 0.521 & 0.800 & 0.983 & 0.276 & 3.500 & 0.769 & 1.000 & 1.000 & 1.000 \\

Collaborator Network Analysis & ReAct & 0.733 & 0.798 & 0.490 & 0.925 & 0.888 & 0.489 & 3.167 & 0.759 & 0.602 & 1.000 & 1.000 \\
 & Onyx & 0.631 & 0.872 & 0.689 & 0.850 & 0.845 & 0.400 & 3.000 & 0.766 & 0.583 & 0.931 & 1.000 \\

Communication Traceback & ReAct & 0.625 & 0.850 & 0.640 & 1.000 & 0.988 & 0.105 & 3.333 & 0.787 & 0.611 & 1.000 & 1.000 \\
 & Onyx & 0.698 & 0.802 & 0.509 & 1.000 & 0.885 & 0.207 & 3.667 & 0.759 & 0.611 & 1.000 & 1.000 \\

Decision And Feedback Recap & ReAct & 0.797 & 0.889 & 0.645 & 0.933 & 0.985 & 0.170 & 3.500 & 0.844 & 0.434 & 1.000 & 1.000 \\
 & Onyx & 0.689 & 0.941 & 0.635 & 0.833 & 0.985 & 0.222 & 3.500 & 0.815 & 0.722 & 0.933 & 1.000 \\

Meeting Prep & ReAct & 0.953 & 0.954 & 0.526 & 0.967 & 0.877 & 0.700 & 3.500 & 0.879 & 0.292 & 1.000 & 1.000 \\
 & Onyx & 0.665 & 0.780 & 0.484 & 0.900 & 1.000 & 0.560 & 2.500 & 0.746 & 0.333 & 0.812 & 1.000 \\

Meeting Recap & ReAct & 0.834 & 0.991 & 0.688 & 0.900 & 0.979 & 0.263 & 4.000 & 0.887 & 0.300 & 1.000 & 1.000 \\
 & Onyx & 0.366 & 0.893 & 0.475 & 0.700 & 0.956 & 0.304 & 2.000 & 0.662 & 0.600 & 0.900 & 1.000 \\

Multi Domain Catch Up & ReAct & 0.760 & 0.918 & 0.567 & 0.971 & 0.940 & 0.531 & 3.000 & 0.827 & 0.375 & 1.000 & 1.000 \\
 & Onyx & 0.597 & 0.838 & 0.709 & 0.914 & 0.955 & 0.417 & 3.286 & 0.772 & 0.524 & 0.946 & 1.000 \\

Multi Hop People Queries & ReAct & 0.394 & 0.805 & 0.651 & 1.000 & 0.918 & 0.296 & 2.625 & 0.695 & 0.563 & 1.000 & 1.000 \\
 & Onyx & 0.413 & 0.733 & 0.794 & 0.875 & 0.923 & 0.271 & 2.750 & 0.689 & 0.729 & 0.958 & 1.000 \\

Project Reflection & ReAct & 0.919 & 0.802 & 0.639 & 0.900 & 0.959 & 0.617 & 3.500 & 0.846 & 0.366 & 1.000 & 1.000 \\
 & Onyx & 0.826 & 0.883 & 0.856 & 0.850 & 0.973 & 0.523 & 3.750 & 0.872 & 0.475 & 0.958 & 1.000 \\

Review Prep & ReAct & 0.824 & 0.856 & 0.533 & 1.000 & 0.913 & 0.299 & 3.333 & 0.821 & 0.375 & 1.000 & 1.000 \\
 & Onyx & 0.766 & 0.838 & 0.671 & 0.960 & 0.890 & 0.457 & 3.600 & 0.811 & 0.625 & 1.000 & 1.000 \\

Time And Collaboration Insights & ReAct & 0.754 & 0.847 & 0.672 & 0.925 & 0.906 & 0.290 & 3.167 & 0.810 & 0.199 & 0.900 & 1.000 \\
 & Onyx & 0.500 & 0.768 & 0.786 & 0.900 & 0.943 & 0.295 & 3.125 & 0.730 & 0.514 & 0.870 & 1.000 \\

\bottomrule
\end{tabular}
\caption{Per-query-type evaluation results across answer quality, retrieval, and ranking metrics.}
\label{tab:query-type-full-results}
\end{table*}

The results in the Table \ref{tab:dataset-results} and \ref{tab:answer-form-comparison} include the metrics computed with gpt-5.4 as a model. However this makes the data generation pipeline, the agent and the metric computation have all the same model. This undermines the reliability of the agent and metric as the same model is essentially judging itself.

We have more experiments where we have used other model families (Grok/Kimi) for both the agent and the evaluator. The agent model is \texttt{Grok 4.1 Fast (reasoning)} (version: \texttt{grok-4-1-fast-reasoning\_1}, API version: \texttt{2025-04-01-preview}), while the judge model is \texttt{Kimi K2.6} (version: \texttt{Kimi-K2.6\_2026-04-20}, API version: \texttt{2025-04-01-preview}). Each experiment is run with a distractor and non-distractor variant. The distractor variant includes the additional distractor emails, which are used to make retrieval more difficult for the answering system.
These additional experiments show a similar (or greater) difficulty in answering the questions when the data generation, the agent and the evaluator all have different models (model families). Results from the experiments are included below.

\begin{table*}[t]
\centering

\label{tab:final-results-onyx}
\resizebox{\linewidth}{!}{%
\begin{tabular}{lcccccccc}
\toprule
\textbf{Metric} &
\multicolumn{2}{c}{\textbf{Cloud}} &
\multicolumn{2}{c}{\textbf{Data}} &
\multicolumn{2}{c}{\textbf{Stripe}} &
\multicolumn{2}{c}{\textbf{VSCode}} \\
\cmidrule(lr){2-3}
\cmidrule(lr){4-5}
\cmidrule(lr){6-7}
\cmidrule(lr){8-9}
&
\textbf{Distr.} &
\textbf{No Distr.} &
\textbf{Distr.} &
\textbf{No Distr.} &
\textbf{Distr.} &
\textbf{No Distr.} &
\textbf{Distr.} &
\textbf{No Distr.} \\
\midrule
final\_score\_avg    & 0.457 & 0.493 & 0.469 & 0.515 & 0.480 & 0.493 & 0.553 & 0.533 \\
completeness         & 0.155 & 0.172 & 0.111 & 0.130 & 0.098 & 0.124 & 0.159 & 0.178 \\
soundness            & 0.536 & 0.573 & 0.448 & 0.535 & 0.511 & 0.554 & 0.650 & 0.588 \\
relevance            & 0.780 & 0.731 & 0.896 & 0.874 & 0.850 & 0.802 & 0.834 & 0.890 \\
faithfulness         & 0.345 & 0.537 & 0.562 & 0.725 & 0.678 & 0.592 & 0.685 & 0.548 \\
readability          & 0.808 & 0.796 & 0.822 & 0.756 & 0.678 & 0.809 & 0.829 & 0.879 \\
retrieval\_accuracy  & 0.353 & 0.464 & 0.409 & 0.444 & 0.486 & 0.619 & 0.324 & 0.544 \\
mrr                  & 0.053 & 0.073 & 0.013 & 0.034 & 0.113 & 0.142 & 0.003 & 0.042 \\
hit\_rate@5          & 0.104 & 0.167 & 0.022 & 0.044 & 0.119 & 0.149 & 0.000 & 0.083 \\
recall@10            & 0.046 & 0.111 & 0.022 & 0.033 & 0.060 & 0.103 & 0.000 & 0.064 \\
citation\_recall\_all & 0.090 & 0.282 & 0.027 & 0.160 & 0.136 & 0.224 & 0.010 & 0.089 \\
\bottomrule
\end{tabular}%
}
\caption{Evaluation results for the Onyx Agent. The evaluations include both distractor and non-distractor variants. It highlights the efficacy of distractor emails in making the dataset challenging.}
\end{table*}

\begin{table*}[t]
\centering
\resizebox{\linewidth}{!}{%
\begin{tabular}{lcccccccc}
\toprule
\textbf{Metric} &
\multicolumn{2}{c}{\textbf{Cloud}} &
\multicolumn{2}{c}{\textbf{Data}} &
\multicolumn{2}{c}{\textbf{Stripe}} &
\multicolumn{2}{c}{\textbf{VSCode}} \\
\cmidrule(lr){2-3}
\cmidrule(lr){4-5}
\cmidrule(lr){6-7}
\cmidrule(lr){8-9}
&
\textbf{Distr.} & \textbf{No Distr.} &
\textbf{Distr.} & \textbf{No Distr.} &
\textbf{Distr.} & \textbf{No Distr.} &
\textbf{Distr.} & \textbf{No Distr.} \\
\midrule
final\_score\_avg      & 0.673 & 0.689 & 0.589 & 0.630 & 0.655 & 0.693 & 0.602 & 0.625 \\
completeness           & 0.282 & 0.334 & 0.147 & 0.165 & 0.231 & 0.283 & 0.141 & 0.175 \\
soundness              & 0.823 & 0.860 & 0.750 & 0.836 & 0.815 & 0.850 & 0.793 & 0.810 \\
relevance              & 0.818 & 0.776 & 0.724 & 0.829 & 0.844 & 0.878 & 0.687 & 0.785 \\
faithfulness           & 0.907 & 0.880 & 0.862 & 0.913 & 0.904 & 0.932 & 0.918 & 0.928 \\
readability            & 0.829 & 0.817 & 0.822 & 0.684 & 0.791 & 0.812 & 0.817 & 0.725 \\
retrieval\_accuracy    & 0.385 & 0.494 & 0.304 & 0.390 & 0.545 & 0.588 & 0.283 & 0.421 \\
mrr                    & 0.324 & 0.385 & 0.221 & 0.382 & 0.275 & 0.462 & 0.176 & 0.258 \\
hit\_rate@5            & 0.542 & 0.667 & 0.444 & 0.511 & 0.448 & 0.507 & 0.312 & 0.438 \\
recall@10              & 0.394 & 0.619 & 0.166 & 0.410 & 0.173 & 0.297 & 0.183 & 0.346 \\
citation\_recall\_all  & 0.599 & 0.807 & 0.357 & 0.570 & 0.330 & 0.600 & 0.358 & 0.576 \\
\bottomrule
\end{tabular}%
}
\caption{Evaluation results for the ReAct Agent. The evaluations include both distractor and non-distractor variants. It highlights the efficacy of distractor emails in making the dataset challenging.}
\label{tab:results}
\end{table*}

Epics and their constituent tasks form dependency graphs, with each stage undergoing structural validation and checks before generating downstream artifacts such as daily diaries. The generation cost scales linearly with the timeline but grows approximately quadratically with the number of employees, epics, and tasks, which are user-defined parameters. Consequently, generating larger datasets becomes prohibitively expensive. Each LLM call averages $\sim$1,000 input and output tokens, with the total number of API calls shown below: Stripe (8,188), VS Code (5,646), Data Platform (5,284), and Cloud Platform (6,129).

\section{Dataset Qualitative Analysis}
\label{sec:detailedexample}

\ameyil{Pick one or two deep examples from the dataset and trace their answer reasoning over emails. We must maybe compare with other datasets slightly on how ours is richer and deeper. I believve this section is important for the story of the paper}

\amey{Nowhere we've clearly defined how the dataset it, what else can we do with the data, like use for further data generation, fine-tuning, building on top of this. For this we must very clearly mention what the data entails, formats, types etc}

In this section we pick a sample QA pair from our dataset, have a walkthrough of the oracle references. The example illustrates a class of questions in our dataset that cannot be
answered by just retrieving a single or even multiple artifact. The system needs to retrieve the right sections from the appropriate set of emails and needs to reason deeply over it to reach a good answer. The asker requests the
\emph{current status} of an in-flight workstream, but \textit{status} is only
meaningful relative to a multi-step protocol that was locked down in earlier
planning artifacts. The question, the expected answer, and the supporting
email evidence are shown across the boxes below, interleaved with
commentary on what makes each part hard.

\begin{center}
\begin{flexprompt}[Q3 Multi Domain Catch Up; 2025-07-28]{promptgrey}{promptgreyborder}
Question (asked by Rina Patel):
Help me work out the current status of the metric-definition reconciliation
workstream.

Answer:
The workstream is still in the orchestration phase, but the foundation is in
place. Earlier planning locked the full reconciliation sequence: start batch
historical recomputation, keep live processing running under the old
definition, buffer incoming events, patch corrected history, replay buffered
events, and only then cut over, with cleanup happening after successful
post-cutover health checks. That planning also finalized the rule that the
Dashboard must stay responsive and explicitly show background progress
without exposing mixed-definition results.

As of today, the workflow coordinator has a first-pass lifecycle model. The
state machine now covers request acceptance, recomputation submitted and
running checkpoints, concurrent old-definition processing, buffered-event
capture, ready-to-patch, patch-and-replay running, ready-to-cutover,
completed, failed, and cleanup-pending states. The coordinator boundaries
are also clearer: recomputation launch, old-definition authority,
buffered-event retention, patch authorization, replay execution, and cutover
readiness are being tracked as explicit phases instead of inferred from
downstream systems.

The main open work is to finalize the transition table and entry and exit
criteria for each state, complete the coordinator rules for concurrent
old-definition processing and buffered-event retention so the streaming path
can implement against a stable contract, and finish the status propagation
and failure semantics needed by patching and cutover consumers. Two
ambiguities that could have blocked this were already addressed:
recomputation launch now uses separate submitted and running checkpoints for
long-running jobs, and failure handling is now modeled as a distinct failed
status with a retry-eligible flag so continuity consumers can show a stable
non-mixed state while recovery remains possible.
\end{flexprompt}
\captionof{figure}{Question and gold answer for a procedural-reasoning
status query. The answer must integrate three distinct layers of truth:
locked invariants from kickoff artifacts, the first-pass coordinator
lifecycle implemented mid-workstream, and the explicitly remaining open
work.}
\label{fig:prompt-orange-reconciliation-qa}
\end{center}

\paragraph{Procedural reasoning over a locked protocol.}
The question looks like a routine status update, but it cannot be answered
by quoting a single email. The model must first reconstruct the locked
reconciliation sequence (recompute $\rightarrow$ old definition live
$\rightarrow$ buffer $\rightarrow$ patch $\rightarrow$ replay
$\rightarrow$ cutover $\rightarrow$ cleanup) and the safety invariants
attached to it (no mixed-definition outputs; cutover gated on replay
completion; cleanup only after post-cutover health checks). These
invariants are established in the kickoff handoff thread, and the gold
answer's first paragraph is essentially a faithful compression of them.

\begin{center}
\begin{flexprompt}[Citations: Kickoff Invariants; 2025-07-22]{promptorange}{promptorangeborder}
  - This is the definitive record for how reconciliation moves from
  definition change through cutover and controlled cleanup, including the
  rule that cutover cannot begin until corrected historical state has been
  applied and buffered events have been replayed without exposing
  mixed-definition results.''
  - `Arjun finished the sequencing notes for old-definition live
  processing, bounded buffering, replay ordering, and cleanup conditions  ... cleanup is recorded as a separate post-cutover activity,
  authorized only after cutover health checks confirm no mixed-definition
  outputs remain visible.
  -`This captures the continuity expectations that keep the Dashboard
  responsive and distinguish background reconciliation from final cutover
  without surfacing inconsistent analytics.
\end{flexprompt}
\captionof{figure}{Kickoff stage email sentences that lock the
reconciliation protocol and its safety invariants. These are the source of
the answer's \textit{locked sequence} and \textit{Dashboard continuity} claims.}
\label{fig:prompt-orange-reconciliation-kickoff}
\end{center}

\paragraph{Separating the \textit{planned} from \textit{implemented}}
A common failure mode for retrieval-augmented systems is to read the
kickoff plan and conclude the workstream is further along than it actually
is. The gold answer carefully distinguishes the locked plan from the
\emph{first-pass} coordinator lifecycle that has actually been built:
explicit lifecycle states, an explicit coordinator boundary, and an
explicit submitted against running split for long-running recomputation. This
distinction is load-bearing without it, the answer would falsely imply
the system is ready to cut over, when in fact the state machine is still a
design artifact.

\begin{center}
\begin{flexprompt}[Citations: Orchestration Update; 2025-07-25)]{promptorange}{promptorangeborder}
  - I translated the agreed stage sequence into a first-pass workflow
  state machine with request accepted, recomputation submitted,
  recomputation running, concurrent old-definition processing active,
  buffered-event capture active, ready-to-patch, patch-and-replay running,
  ready-to-cutover, completed, failed, and cleanup-pending states.
  - I also drafted the coordinator split so we can launch historical
  recomputation from the epic 3 trigger inputs while keeping a separate
  lifecycle record for authorization and recovery decisions instead of
  inferring everything from downstream job state.
  - We also closed the recomputation acknowledgment gap by splitting
  acceptance from execution confirmation, so the coordinator can record
  submitted immediately and wait for a later running checkpoint once
  downstream recomputation actually starts.
\end{flexprompt}
\captionof{figure}{Mid-workstream sentences that establish what has
\emph{actually} been implemented: a first-pass state machine, an explicit
coordinator boundary, and the submitted-vs-running checkpoint split.}
\label{fig:prompt-orange-reconciliation-orch}
\end{center}

\paragraph{Resolved ambiguities and remaining open work.}
The hardest part of the answer is the final paragraph is identifying which
ambiguities have already been closed (so they should not appear as open
risks) and which contracts are still missing (so downstream consumers
cannot yet build against them). This requires reading replies in the
thread, not just the lead message the streaming side response is what
elevates buffered-event retention from a streaming assumption to an
explicit coordinator rule, and what frames the patch/replay transition as
gated on recorded prerequisites rather than downstream timing. A model
that summarizes only the most recent message will silently drop these
constraints.

\begin{center}
\begin{flexprompt}[{Citations: Streaming Side Reply; 2025-07-25}]{promptorange}{promptorangeborder}
  - old-definition processing should remain authoritative until
  replay prerequisites are satisfied, not just until recomputation is
  marked running.
  - buffered-event retention needs its own coordinator-side rule so
  T4 has a stable contract for when replayable data must still be
  preserved.
  - transitions into the patch/replay path should stay blocked until
  the replay prerequisites are actually recorded, rather than inferred
  from downstream timing.
  - I'm aligned with the submitted vs running split for recomputation
  acknowledgment that gives us a cleaner separation between launch
  acceptance and long-running execution state.
\end{flexprompt}
\captionof{figure}{Streaming side reply sentences that pin down the
remaining open coordinator contracts (old definition authority,
buffered-event retention, patch/replay gating) and confirm the resolved
submitted instead of running ambiguity.}
\label{fig:prompt-orange-reconciliation-stream}
\end{center}

\paragraph{Why this example is difficult overall?}

The question is a strong test of multi document, cross thread reasoning
under safety constraints. The model must (i)~reconstruct an ordered
protocol from kickoff artifacts, (ii)~separately read the orchestration
update to identify the first-pass lifecycle that exists today,
(iii)~incorporate a reply from another author to determine which
coordinator rules are still open, and (iv)~preserve hard invariants
throughout (no mixed definition outputs; cutover gated on replay; cleanup
gated on health checks). A surface level \textit{latest update} summary
produces a fluent but incorrect answer, it tends to either overstate
progress or omit the specific resolved ambiguities (submitted or running
checkpoints; failed status with a \texttt{retry\_eligible} flag) that distinguish a
genuine status assessment from a generic project recap.

\section{Dataset Examples}
\label{app:C}
\subsection{Prompts}

In this section we include the actual prompts used in our pipeline at various stages. 

\subsubsection{Data Generation Prompts}

\begin{center}
\begin{flexprompt}[Epics]{promptgreen}{promptgreenborder}
TASK:
Extract a structured list of all epics, projects, and initiatives from the newsletter.

INSTRUCTIONS:
- Extract a comprehensive list of epics mentioned or implied by the newsletter, ensuring realistic scope and sufficient detail.
- Epics should collectively form a coherent plan leading to the outcomes described.
- Assign start_date/end_date within the provided timeline window.
- Assign 3-5 employees per epic using the provided employee list and reporting structure.
- Use predecessor_epics to form a DAG; avoid cycles.
- Output must strictly match the JSON template.

NEWSLETTER CONTENT:
{newsletter_content}

EMPLOYEES:
{employees}

REPORTING STRUCTURE:
{reporting_structure}

TIMELINE WINDOW:
- start: {timeline_start}
- end: {timeline_end}

OUTPUT TEMPLATE:
{output_template}

OUTPUT:
\end{flexprompt}
\captionof{figure}{Prompt used for generating \textit{Epics}}
\label{fig:promptepic}
\end{center}

\begin{center}
\begin{flexprompt}[Tasks]{promptgreen}{promptgreenborder}
TASK:
You are an expert at decomposing a project epic into concrete tasks and producing a direct dependency graph (DAG).

GOAL:
Given a epic, produce a set of tasks T1..Tn that cover the full epic scope and form an explicit DAG via depends_on_tasks.
This graph will be used as the single source of truth for generating daily breakdowns (we are skipping weekly breakdown generation).

INSTRUCTIONS:
1. Create between 6 and 20 tasks (T1..Tn), sized so they can be progressed across multiple business days.
    - T1 MUST be a discussion/high-level planning and handoff task (kickoff, scope alignment, predecessor review, decision log, or stakeholder alignment) that sets up the rest of the epic.
2. Every task must have:
    - id: "Tk" format, unique, sequential starting at T1
    - title, description, expected_outputs, unlocks
    - assigned_employees: list of 1+ employee names, each must be one of the assigned epic employees
    - timeline: start_date/end_date/business_dates within the epic window
3. Dependencies:
    - Use depends_on_tasks to represent direct prerequisites among T-items.
    - The dependency structure MUST be a DAG (no cycles).
    - Only reference IDs that exist in this output.
4. Timeline correctness (CRITICAL):
    - timeline MUST be consistent with the DAG: if T2 depends_on_tasks includes T1, then T2.timeline.start_date MUST be on or after T1.timeline.end_date.
    - business_dates MUST be weekdays only and fall within [start_date, end_date].
4. Predecessor/successor awareness:
    - If predecessor epics provide required inputs/decisions, capture them in depends_on_predecessor_deliverables.
    - If successor epics need handoffs, make those handoffs explicit in expected_outputs/unlocks.
6. Use newsletter segment content to ground objectives, but treat any epic.newsletter_segments entries with label="future_reference" as CONTEXT-ONLY:
     - You MAY use them to understand what the work is ultimately driving toward.
     - You MUST NOT mention, quote, paraphrase, or otherwise reveal their content in the output.
     - You MUST NOT incorporate future outcomes/metrics as if they are already known during the epic.

CONTEXT:

epic details:
{epic}

Predecessor epics (dependencies):
{predecessor_epics}

Successor epics (downstream epics):
{successor_epics}

Newsletter content (for context):
{newsletter_content}

Assigned employee details (ONLY assigned to this epic; do not use or invent other employee profiles):
{epic_employees}

Output template:
{task_graph_template}

OUTPUT:
- Return JSON matching the template exactly.
- Do not include any explanatory text, commentary, reasoning, or metadata before or after the output
- Do not include code fences, ```json tags, or any markdown formatting
 - Any free-text fields (titles, descriptions, expected_outputs, unlocks) must NOT mention the words "newsletter", "article", "document", or otherwise refer to the source text; they should describe the work itself.
\end{flexprompt}
\captionof{figure}{Prompt used for generating \textit{Tasks}}
\label{fig:prompttask}
\end{center}

\amey{Add some textual description on what the prompts do}

\subsubsection{Validation Reflection Refinement Prompts}

\begin{center}
\begin{flexprompt}[Email Validation]{promptgrey}{promptgreyborder}
TASK:
You are an expert at validating email conversations for consistency, logical flow, and factual accuracy against source materials.

GOAL:
Validate generated or existing email conversations for authenticity, consistency with extracted facts, proper timeline sequencing, realistic communication patterns, and adherence to organizational context. Identify gaps, anomalies, inconsistencies, or logical issues in email threads.

INSTRUCTIONS:
1. Thoroughly review the complete email conversation thread
2. Validate each email against the following criteria:
    - Temporal consistency (dates in correct chronological order)
    - Participant consistency (senders/recipients match organizational roles and previous communications)
    - Content consistency (no contradictions with previously stated information)
    - Factual accuracy (all facts referenced match extracted facts or known information)
    - epic progression (logical flow toward epic objectives)
    - Tone and style consistency (professional tone maintained throughout)
    - Format compliance (follows email template structure)
3. Cross-reference email content with extracted facts to identify:
    - Facts that are missing from email discussions
    - Facts mentioned in emails that contradict source material
    - Email content not covered by extracted facts
    - Logical gaps or unexplained transitions between emails
4. Flag specific issues including:
    - Temporal inconsistencies (out-of-order dates or illogical timing)
    - Logical gaps (missing information or unexplained transitions)
    - Factual errors or contradictions (conflicts with extracted facts)
    - Role inconsistencies (sender/recipient mismatches)
    - Uncovered content (information in emails not present in facts)
    - Missing coverage (facts not discussed in emails)
5. Provide a concise summary highlighting:
    - What is missing from emails relative to extracted facts
    - What is incoherent or contradictory within the email thread
    - What email content is not covered by the extracted facts

CONTEXT:
Email:
{email}

Extracted facts from newsletter corresponding to the email:
{fact_mapping}

Output Validation Template (specifying required fields, data types, and expected format):
{email_validation_template}

OUTPUT:
- Generate a structured validation report exactly matching the provided validation template format.
- Include a concise summary of gaps, inconsistencies, contradictions, and uncovered content.
- Adhere exactly to the structure specified in Output JSON Template, use double quotes for JSON keys and string values.
- Do not include any explanatory text, commentary, reasoning, or metadata before or after the output
- Do not include code fences, ```json tags, or any markdown formatting
\end{flexprompt}
\captionof{figure}{Prompt used for validating the \textit{Emails}}
\label{fig:promptemailvalidation}
\end{center}

\begin{center}
\begin{flexprompt}[Email Reflection]{promptgrey}{promptgreyborder}
TASK:
You are an expert at validating email conversations for consistency, logical flow, and epic deliverable achievement.

GOAL:
Validate generated or existing email conversations for authenticity, consistency with epic deliverables, proper timeline sequencing, realistic communication patterns, and adherence to organizational context. Identify whether all deliverables were addressed and achieved throughout the email thread, and flag any gaps, anomalies, inconsistencies, or logical issues.

INSTRUCTIONS:
1. Thoroughly review the complete email conversation thread
2. Validate each email against the following criteria:
    - Temporal consistency (dates in correct chronological order)
    - Participant consistency (senders/recipients match organizational roles and previous communications)
    - Content consistency (no contradictions with previously stated information)
    - Deliverable progress (emails show advancement toward epic deliverables)
    - epic progression (logical flow toward epic objectives)
    - Tone and style consistency (professional tone maintained throughout)
    - Format compliance (follows email template structure)
3. Cross-reference email content with epic deliverables to identify:
    - Deliverables that are missing from email discussions
    - Deliverables mentioned in emails with evidence of completion or progress
    - Email content not aligned with any deliverable
    - Logical gaps or unexplained transitions between emails
4. For each deliverable, determine:
    - Whether it was addressed in the email thread
    - Level of completion (not started, in progress, completed)
    - Supporting evidence from specific emails
5. Flag specific issues including:
    - Temporal inconsistencies (out-of-order dates or illogical timing)
    - Logical gaps (missing information or unexplained transitions)
    - Unaddressed deliverables (deliverables not discussed in emails)
    - Role inconsistencies (sender/recipient mismatches)
    - Uncovered content (information in emails not related to any deliverable)
6. Provide a concise summary highlighting:
    - Which deliverables were achieved in the email thread
    - Which deliverables remain unaddressed or incomplete
    - What is incoherent or contradictory within the email thread
    - Critical gaps between epic requirements and email evidence

CONTEXT:
Email:
{email_conversations}

epic Deliverables:
{deliverables}

Output Validation Template (specifying required fields, data types, and expected format):
{email_validation_template}

OUTPUT:
- Generate a structured validation report exactly matching the provided validation template format.
- Include a concise summary of deliverable achievement status, gaps, inconsistencies, and unaddressed requirements.
- Adhere exactly to the structure specified in Output JSON Template, use double quotes for JSON keys and string values.
- Do not include any explanatory text, commentary, reasoning, or metadata before or after the output
- Do not include code fences, ```json tags, or any markdown formatting
\end{flexprompt}
\captionof{figure}{Prompt used for reflecting on the \textit{Emails}}
\label{fig:promptemailreflection}
\end{center}

\subsection{Evaluation Prompts}
\label{app:eval-prompts}

We use structured prompting templates to evaluate system outputs across multiple dimensions. Each prompt enforces explicit reasoning steps and a JSON output schema to ensure consistency and interpretability.

\subsubsection{Completeness}
\textbf{Inputs:} Question, Reference Answer, Candidate Answer \\
\textbf{Output:} JSON with score in $[0,1]$

\begin{center}
\begin{flexprompt}[Completeness]{promptblue}{promptblueborder}
TASK:
Evaluate the completeness of a candidate answer.

INSTRUCTIONS:
- Identify each distinct information point in the reference answer.
- Check whether each information point is present in the candidate answer.
- Count paraphrased or equivalent information as covered.
- Penalize missing reference information proportionally.
- Do not penalize additional correct information beyond the reference.
- Ignore formatting and style.

SCORING:
score = covered_reference_points / total_reference_points

INPUTS:
Question:
{question}

Reference Answer:
{reference}

Candidate Answer:
{candidate}

OUTPUT:
Return JSON only:
{
  "reference_points": [
    "list each distinct information point from the reference"
  ],
  "covered_points": [
    "reference points present in the candidate"
  ],
  "missing_points": [
    "reference points missing from the candidate"
  ],
  "score": 0.0,
  "reasoning": "brief justification"
}
\end{flexprompt}
\captionof{figure}{Prompt used for completeness evaluation.}
\label{fig:prompt-completeness}
\end{center}

\subsubsection{Soundness}
\textbf{Inputs:} Question, Reference Answer, Candidate Answer \\
\textbf{Output:} JSON with score in $[0,1]$

\begin{center}
\begin{flexprompt}[Soundness]{promptblue}{promptblueborder}
TASK:
Evaluate the factual correctness of a candidate answer.

INSTRUCTIONS:
- Identify each factual assertion in the candidate answer.
- Mark each assertion as correct, incorrect, or unverifiable.
- Mark an assertion incorrect if it contradicts the reference answer.
- Mark an assertion incorrect if it attributes information to the wrong entity, document, or date.
- Do not penalize additional correct information beyond the reference.
- An empty answer or "I don't know" is not unsound, but it may be incomplete.

SCORING:
score = 1 - (num_incorrect / total_assertions)

INPUTS:
Question:
{question}

Reference Answer:
{reference}

Candidate Answer:
{candidate}

OUTPUT:
Return JSON only:
{
  "assertions": [
    {
      "assertion": "text of the claim",
      "verdict": "correct|incorrect|unverifiable",
      "reason": "why"
    }
  ],
  "num_correct": 0,
  "num_incorrect": 0,
  "num_unverifiable": 0,
  "score": 0.0,
  "reasoning": "brief overall justification"
}
\end{flexprompt}
\captionof{figure}{Prompt used for soundness evaluation.}
\label{fig:prompt-soundness}
\end{center}

\subsubsection{Relevance}
\textbf{Inputs:} Question, Reference Answer, Candidate Answer \\
\textbf{Output:} JSON with score in $[0,1]$

\begin{center}
\begin{flexprompt}[Relevance]{promptblue}{promptblueborder}
TASK:
Evaluate whether the candidate answer stays focused on the question.

INSTRUCTIONS:
- Identify each distinct statement in the candidate answer.
- Mark each statement as relevant or superfluous.
- Relevant statements directly help answer the question.
- Superfluous statements may be correct but are not needed to answer the question.
- Brief contextual framing is acceptable.
- Penalize lengthy digressions, unrelated facts, or excessive background.
- Empty answers or "I don't know" are trivially relevant but may be incomplete.

SCORING:
score = num_relevant / total_statements

INPUTS:
Question:
{question}

Reference Answer:
{reference}

Candidate Answer:
{candidate}

OUTPUT:
Return JSON only:
{
  "statements": [
    {
      "statement": "text",
      "classification": "relevant|superfluous",
      "reason": "why"
    }
  ],
  "num_relevant": 0,
  "num_superfluous": 0,
  "score": 0.0,
  "reasoning": "brief overall justification"
}
\end{flexprompt}
\captionof{figure}{Prompt used for relevance evaluation.}
\label{fig:prompt-relevance}
\end{center}

\subsubsection{Readability}
\textbf{Inputs:} question, candidate answer \\
\textbf{Output:} JSON with readability score on a 0--5 scale

\begin{center}
\begin{flexprompt}[Readability]{promptblue}{promptblueborder}
TASK:
Evaluate readability and format adherence of a candidate answer.

INSTRUCTIONS:
- Identify whether the question requests a specific output format.
- Score readability on a 0-5 scale based on clarity, structure, and ease of parsing.
- Score format adherence on a 0-5 scale if a format is requested.
- Use null for format_adherence_score if no format is requested.
- Check for residual reasoning, unnecessary repetition, or garbled text.
- Consider whether lists, IDs, links, or tables are presented in a scannable way.

SCORING:
1 = exceptionally clear and well organized.
0.8 = clear and well organized with minor issues.
0.6 = understandable but could be better organized.
0.4 = somewhat confusing and requires effort to parse.
0.2 = poorly organized or hard to follow.
0 = incomprehensible.

INPUTS:
Question:
{question}

Candidate Answer:
{candidate}

OUTPUT:
Return JSON only:
{
  "format_requested": "description of requested format, or null",
  "readability_score": 0,
  "readability_issues": [
    "specific issues, if any"
  ],
  "format_adherence_score": 0,
  "format_adherence_issues": [
    "specific issues, if any"
  ],
  "has_residual_reasoning": false,
  "reasoning": "brief overall justification"
}
\end{flexprompt}
\captionof{figure}{Prompt used for readability evaluation.}
\label{fig:prompt-readability}
\end{center}

\subsubsection{Retrieval Accuracy}
\textbf{Inputs:} question, retrieved chunks, gold attribution documents \\
\textbf{Output:} JSON with chunk precision and document recall

\begin{center}
\begin{flexprompt}[Retrieval Accuracy]{promptblue}{promptblueborder}
TASK:
Evaluate retrieval accuracy for a RAG system.

INSTRUCTIONS:
- For each retrieved chunk, decide whether it is relevant to answering the question.
- A chunk is relevant if it contains information that could help answer the question, even partially.
- A chunk is irrelevant if it contains no useful information for the question.
- If gold attribution documents are provided, determine which gold documents are covered by at least one retrieved chunk.
- A retrieved chunk covers a gold document if it contains information from that gold document, even if chunk boundaries differ.

SCORING:
chunk_precision = num_relevant_chunks / total_retrieved_chunks
document_recall = num_covered_gold_documents / total_gold_documents

INPUTS:
Question:
{question}

Retrieved Chunks:
{evidence_block}

Gold Attribution Documents:
{gold_docs_block}

OUTPUT:
Return JSON only:
{
  "chunk_assessments": [
    {
      "chunk_id": "identifier",
      "relevant": true,
      "reason": "brief justification"
    }
  ],
  "num_relevant_chunks": 0,
  "num_irrelevant_chunks": 0,
  "chunk_precision": 0.0,
  "gold_document_coverage": [
    {
      "gold_doc_id": "identifier",
      "covered": true,
      "covered_by_chunk": "chunk_id or null"
    }
  ],
  "document_recall": 0.0,
  "reasoning": "brief overall justification"
}
\end{flexprompt}
\captionof{figure}{Prompt used for retrieval accuracy evaluation.}
\label{fig:prompt-retrieval-accuracy}
\end{center}

\subsubsection{Faithfulness}
\textbf{Inputs:} question, candidate answer, retrieved chunks \\
\textbf{Output:} JSON with score in $[0,1]$

\begin{center}
\begin{flexprompt}[Faithfulness]{promptblue}{promptblueborder}
TASK:
Evaluate whether the candidate answer is grounded in the retrieved evidence.

INSTRUCTIONS:
- Break the candidate answer into individual claims.
- For each claim, search the retrieved chunks for supporting evidence.
- Mark each claim as supported, partially_supported, or unsupported.
- Supported means the retrieved chunks directly support the claim.
- Partially supported means the retrieved chunks contain related evidence but do not fully confirm the claim.
- Unsupported means no retrieved chunk supports the claim.
- Common knowledge or trivially true statements should be marked as supported.
- Faithfulness measures grounding in retrieved evidence, not factual correctness against the reference.

SCORING:
score = (num_supported + 0.5 * num_partially_supported) / total_claims

INPUTS:
Question:
{question}

Candidate Answer:
{candidate}

Retrieved Chunks:
{evidence_block}

OUTPUT:
Return JSON only:
{
  "claims": [
    {
      "claim": "text of the claim",
      "verdict": "supported|partially_supported|unsupported",
      "evidence_chunk_id": "chunk_id that supports this claim, or null",
      "reason": "brief justification"
    }
  ],
  "num_supported": 0,
  "num_partially_supported": 0,
  "num_unsupported": 0,
  "score": 0.0,
  "reasoning": "brief overall justification"
}
\end{flexprompt}
\captionof{figure}{Prompt used for faithfulness evaluation.}
\label{fig:prompt-faithfulness}
\end{center}

\subsubsection{Overall Score}
\textbf{Inputs:} Question, Reference Answer, Candidate Answer, retrieved chunks, gold documents, sub-scores \\
\textbf{Output:} integer score on a 0--5 scale

\begin{center}
\begin{flexprompt}[Overall Score]{promptblue}{promptblueborder}
TASK:
Provide a holistic correctness assessment of a candidate answer for a RAG system.

INSTRUCTIONS:
- Review the Question, Reference Answer, Candidate Answer, retrieved evidence, gold attribution documents, and sub-scores.
- Use the sub-scores as guidance, but apply holistic judgment.
- The overall score is not a mechanical average.
- Penalize answers that are incomplete, misleading, factually incorrect, poorly grounded, irrelevant, or hard to parse.
- If the answer is missing key information from the reference, it cannot score above 3 even if everything it says is correct.
- If the answer contains hallucinations not supported by retrieved documents, penalize proportionally.

SCORING GUIDE:
5 = essentially perfect: complete, correct, grounded, concise, and well formatted.
4 = good with only minor gaps or issues.
3 = acceptable but has notable weaknesses.
2 = major problems, such as missing most key points or significant errors.
1 = poor: mostly wrong, irrelevant, or ungrounded.
0 = empty, incomprehensible, or completely fails to answer.

INPUTS:
Question:
{question}

Reference Answer:
{reference}

Candidate Answer:
{candidate}

Gold Attribution Documents:
{gold_docs_block}

Retrieved Evidence:
{evidence_block}

Sub-scores:
{sub_scores_block}

OUTPUT:
Return JSON only:
{
  "score": 0,
  "reasoning": "brief justification for the overall score"
}
\end{flexprompt}
\captionof{figure}{Prompt used for overall evaluation.}
\label{fig:prompt-overall}
\end{center}

\subsection{Synthetic Data Output}

This section contains actual data generated from our pipeline.

\subsubsection{Intermediate Artifacts}

% \begin{flexprompt}[Title]{promptgrey}{promptgreyborder}
% Body
% \end{flexprompt}

% \begin{flexprompt}[Title]{promptgreen}{promptgreenborder}
% Body
% \end{flexprompt}

% \begin{flexprompt}[Title]{promptblue}{promptblueborder}
% Body
% \end{flexprompt}

% \begin{flexprompt}[Title]{promptorange}{promptorangeborder}
% Body
% \end{flexprompt}

% \begin{flexprompt}[Title]{promptpurple}{promptpurpleborder}
% Body
% \end{flexprompt}

% \begin{flexprompt}[Title]{promptgreen}{promptgreenborder}
% Body
% \end{flexprompt}

\begin{center}
\begin{flexprompt}[Epic]{promptpurple}{promptpurpleborder}
"id": "2",
    "title": "Agent Sessions, Session Search, and Plan-First Workflow Implementation",
    "description": "Deliver the core Agent HQ workflow foundation by implementing the Agent Sessions view as the central place to manage active sessions across local and background agents, including default enablement, source-based organization....",
    "newsletter_segments": [
      {
        "text": "As you hand off tasks to various coding agents, it's important to have a clear overview of all your active sessions.",
        "label": "other"
      },
      {
        "text": "This includes both local sessions in VS Code and sessions created by background agents in other environments, such as Copilot coding agent, GitHub Copilot CLI, or OpenAI's Codex.",
        "label": "other"
      },
      .
      .
      .
    "start_date": "2025-08-18",
    "end_date": "2025-08-29",
    "assigned_employees": [
      "Maya Chen",
      "Elena Morozova",
      "Priya Raman"
    ],
    "predecessor_epics": [
      "1"
    ],
    "dependency_notes": "This work depends on the release-framing epic to align the Agent HQ theme and user-facing session-management goals. It produces the core session overview and plan-approval foundation that later epics rely on for cloud and CLI delegation, deeper custom agent integration, and downstream chat workflow refinements.",
    "assignment_notes": "No assignment changes were required. The existing product and engineering ownership remains well aligned to session management and plan-first workflow delivery.",
    "duration_days": 12
\end{flexprompt}
\captionof{figure}{Example of Epic}
\label{fig:artifactepic}
\end{center}

\begin{center}
\begin{flexprompt}[Task]{promptpurple}{promptpurpleborder}
{
      "id": "T2",
      "title": "Implement default Agent Sessions view structure and placement controls",
      "description": "Build the centralized Agent Sessions view as the primary management surface for active sessions, including default enablement, source-based organization, separate local and background sections, and the configurable placement behavior controlled by the chat.agentSessionsViewLocation setting with support for the optional single-view location beside Chat in the Secondary Side Bar.",
      "assigned_employees": [
        "Elena Morozova",
        "Maya Chen"
      ],
      "timeline": {
        "start_date": "2025-08-22",
        "end_date": "2025-08-25",
        "business_dates": [
          "2025-08-22",
          "2025-08-25"
        ]
      },
      "depends_on_tasks": [
        "T1"
      ],
      "depends_on_predecessor_deliverables": [
        "Agent HQ positioning for a single session overview across local and remote agent activity",
        "Agreed implementation sequence for session view, search, planning flow, and custom plan agent configuration"
        .
        .
        .
      ],
      "expected_outputs": [
        "Centralized Agent Sessions management surface covering active local VS Code sessions and background agent sessions from supported external environments",
        .
        .
        .
      ],
      "unlocks": [
        "Enables search integration against a stable session list surface",
        "Provides the UI and configuration foundation required by later cross-environment session integrations",
        .
        .
        .
      ]
    },
\end{flexprompt}
\captionof{figure}{Example of Task}
\label{fig:artifacttask}
\end{center}

\begin{center}
\begin{flexprompt}[Daily Diary]{promptpurple}{promptpurpleborder}
"daily_entries": [
    {
      "day": "Friday",
      "date": "2025-08-22",
      "day_type": "deep work",
      "work_summary": "Elena Morozova and Maya Chen began the session-surface implementation using the locked T1 decisions, establishing the baseline view structure, default enablement behavior, and the setting model for standard versus consolidated placement.",
      "detailed_activities": [
        {
          "employee": "Elena Morozova",
          "activity": "Elena Morozova reviewed the finalized T1 decision log and translated it into an implementation plan for the Agent Sessions surface ..."
        },
        .
        .
        .
      ],
      "blockers": [
        {
          "description": "While wiring the placement setting, Elena Morozova found that the decision log defined the optional consolidated layout as adjacent to Chat in the Secondary Side Bar ...",
          "reported_by": "Elena Morozova"
        }
      ],
      "resolutions": [
        {
          "description": "Maya Chen resolved the layout ambiguity by documenting that the single-view arrangement may consolidate the surface placement, but it must continue to preserve source-aware organization and must not be treated as feature-equivalent to the standard sectioned layout; ...",
          "resolved_by": "Maya Chen",
          "blocker_reference": "While wiring the placement setting, Elena Morozova found that the decision log defined the optional consolidated layout as adjacent to Chat in the Secondary Side Bar ..."
        }
      ],
      "collaboration_interactions": [
        {
          "participants": [
            "Elena Morozova",
            "Maya Chen"
          ],
          "interaction_type": "design discussion",
          "description": "Elena Morozova walked Maya Chen through the initial view architecture for the centralized sessions surface, covering section separation ....",
          "initiator": "Elena Morozova",
          "primary_participants": [
            "Maya Chen"
          ],
          "secondary_participants": []
        },
        .
        .
      ],
      "progress_notes": [
        {
          "note": "The implementation started on top of the T1 decisions with the baseline structure for a centralized sessions surface and distinct local and background sections in place.",
          "contributors": [
            "Elena Morozova",
            "Maya Chen"
          ]
        },
        .
        .
      ],
      "next_steps": [
        {
          "owner": "Elena Morozova",
          "action": "Complete the setting-backed placement behavior for both standard and single-view layouts and verify adjacency to Chat in the Secondary Side Bar."
        },
        .
        .
      ],
      "references": []
    },
\end{flexprompt}
\captionof{figure}{Example of Daily Diary}
\label{fig:artifactdiary}
\end{center}

\subsubsection{Final Dataset}

\begin{center}
\begin{flexprompt}[Email Thread]{promptorange}{promptorangeborder}
{
        "id": "L25",
        "from": "elena.morozova@microsoft.com",
        "to": [
          "maya.chen@microsoft.com"
        ],
        "cc": [
          "avery.sinclair@microsoft.com",
          "priya.raman@microsoft.com"
        ],
        "timestamp": "2025-08-22T10:17:00",
        "subject": "Update: T2 - Implement default Agent Sessions view structure and placement controls",
        "body":
        "Hi Maya,Quick note to start the thread for T2. I translated the locked T1 decisions into the implementation plan and have the initial Agent Sessions surface wired around active local VS Code sessions plus background agent sessions from supported external environments.I built the standard layout container with separate local and background sections-connected the session list model so items can stay organized by source within those sections .... If you’re aligned, I’ll keep the implementation notes explicit that the sectioned layout is the default experience and track the remaining consolidated-layout differences in the limitations list while I finish the placement behavior.",
        "references_epic_diary": {
          "date": "2025-08-22",
          "summary": "Relates to the initial implementation plan, baseline sectioned view structure, default enablement and placement-setting wiring, plus the layout ambiguity Elena raised during design review."
        }
      },
      {
        "id": "L26",
        "from": "maya.chen@microsoft.com",
        "to": [
          "elena.morozova@microsoft.com"
        ],
        "cc": [
          "avery.sinclair@microsoft.com",
          "priya.raman@microsoft.com"
        ],
        "timestamp": "2025-08-22T11:06:00",
        "subject": "Re: Update: T2 - Implement default Agent Sessions view structure and placement controls",
        "body": "Hi Elena,Yes, that matches the product rule I wanted to lock today.Please treat these as the acceptance boundaries for the current pass: the centralized sessions surface is the baseline experience, the default presentation remains the sectioned layout with distinct local and background sections .... the main thing is to keep adjacency to Chat in the Secondary Side Bar tied to the setting path without weakening the default sectioned experience. Sounds good to continue on that basis.",
        "references_epic_diary": {
          "date": "2025-08-22",
          "summary": "Relates to Maya's follow-up product guidance resolving the single-view ambiguity and defining baseline-versus-consolidated behavior, including explicit limitation handling."
        }
      },
      {
        "id": "L31",
        "from": "maya.chen@microsoft.com",
        "to": [
          "marcus.alvarez@microsoft.com",
          "rohan.mehta@microsoft.com"
        ],
        "cc": [
          "avery.sinclair@microsoft.com",
          "elena.morozova@microsoft.com",
          "priya.raman@microsoft.com"
        ],
        "timestamp": "2025-08-25T14:11:00",
        "subject": "Re: Update: T2 - Implement default Agent Sessions view structure and placement controls",
        "body": "Hi Rohan and Marcus,Following up on the session-surface thread, we wrapped T2 today and I’m handing over the pieces your next phase will build on. .... Elena also fixed the transition-path issue we found this morning so the consolidated placement now keeps the same source-aware grouping metadata needed by later search and integration work.\n\nIf helpful, we can keep using this thread for any contract questions as you start wiring search and cross-environment behaviors on top of the stabilized list surface.",
      },
\end{flexprompt}
\captionof{figure}{Example of an Email Thread}
\label{fig:outputemailthread}
\end{center}

\begin{center}
\begin{flexprompt}[Example of Daily Briefing Question and Answer]{promptorange}{promptorangeborder}
"asked_on": "2025-10-30",
"asked_by": "Leila Haddad",
"question": "What should I focus on at work today?",   
"answer": 

"**Overview** 
    You completed the UX continuity section yesterday during the handoff work and documented the interaction and discoverability constraints that downstream validation work needs to preserve. Based on yesterday's handoff-oriented next steps, your focus today should be on advising downstream teams as release-close polish work begins, especially where interaction continuity or discoverability could drift.

**Key Updates Since Yesterday** 
    The finalized handoff package was completed and published, including the decision log, risk register, owners matrix, and integrated execution plan.
    - You completed the UX continuity section of the handoff review, documenting interaction and discoverability constraints that downstream validation tasks need to preserve while user-facing polish is finalized.
    - Dependency review details were converted into explicit handoff expectations so downstream teams can move forward without reopening earlier scope questions.
    - A cross-workstream escalation ambiguity in the owners matrix was resolved by assigning product-level arbitration through the documented governance path and capturing the operational escalation path in the execution plan.
    - In a final chat thread including Maya Chen, Rohan Mehta, Leila Haddad, and Sofia Lindberg, the group confirmed that the published owners matrix matched the scope notes and that no downstream task would inherit undocumented UX, engineering, or packaging assumptions.
 
 **Suggested Epics for Today** 
    - Prioritize advising downstream teams on preserving interaction and discoverability continuity while they validate release-close polish. That is your explicit next step from the most recent work.
    - Be ready to answer questions that connect user-facing polish to the constraints you documented in the handoff review, since the handoff package is now in use by downstream owners.
 - Watch for scope drift where downstream examples or validation work could expand beyond the intended surface area. Based on recent activity, that is particularly relevant where UI scenario guidance could blur into broader product workflows.
    
**Meeting Prep for Today** - No meetings identified for today.
    
\end{flexprompt}
\captionof{figure}{Example of Daily Briefing Question and Answer}
\label{fig:outputdailybriefingqa}
\end{center}

\begin{center}
\begin{flexprompt}[Example of Project Status Question and Answer]{promptorange}{promptorangeborder}
    "asked_on": "2025-10-30",
    "asked_by": "Avery Sinclair",
    "question": "What is the current status of the complete project?",
    "answer": 
    
    "**Project Overview**
     The project has progressed from release framing through delivery of session management, trust controls, terminal runtime work, and chat/editor experience updates; the current focus is release-close stabilization, API packaging, ecosystem validation, and operational follow-through.
     - Scope now spans shipped platform polish, remaining developer-facing APIs and ecosystem guidance, plus final validation of source control, testing, language, and extension behaviors. Based on recent activity, the project is in closure and stabilization rather than core feature build-out.
     
     **Key Updates**
     - 2025-10-24: Sofia Lindberg, Jonas Richter, Camila Torres, and Victor Nguyen closed the chat/editor experience phase, completed the validated acceptance checklist, packaged deliverables, known-issue log, ownership map, and formally transferred stabilization-ready artifacts to the release-tail owners.
     - 2025-10-23: Camila Torres, Victor Nguyen, Nadia Petrov, and Jonas Richter completed integrated cross-surface validation, confirmed sign-off readiness across prompt workflows, reasoning, model management, accessibility-sensitive defaults, and editor improvements, and produced a triaged issue list with accepted limitations and watch areas.
     - 2025-10-28: Ethan Wallace, Avery Sinclair, Rohan Mehta, Leila Haddad, Maya Chen, and Sofia Lindberg finalized the release-close kickoff package, including the decision log, owners matrix, risk register, and integrated execution plan for downstream stabilization, ecosystem, API, and operational tracks.
     - 2025-10-29: Rohan Mehta, Grace Okafor, Sofia Lindberg, and Noah Patel started the extension authoring API packaging work and drafted the completion pack, compatibility notes, scenario guide, and rollout guidance structure.
     
     **Risks & Blockers**
     - Medium: The API packaging work initially blurred stable contracts and proposed API scenarios, which could cause partners to misread proposal guidance as committed behavior. Mitigation: Noah Patel restructured the compatibility document into stable contracts, proposed scenarios, and consumer caveats on 2025-10-29.
     - Medium: Release-close kickoff surfaced cross-workstream ambiguity between API packaging and ecosystem validation. Mitigation: Avery Sinclair and Rohan Mehta split responsibilities into API contract/compatibility packaging versus ecosystem behavior validation and communication consistency on 2025-10-27.
     - Low: The owners matrix initially left arbitration unclear for issues spanning source control polish and ecosystem packaging. Mitigation: Avery Sinclair assigned product-level arbitration and Ethan Wallace documented the operational escalation path on 2025-10-28.
     
     **Next Steps & Upcoming Milestones**

    - Release-close execution plan follow-through
      Owner: Ethan Wallace
      Status: In Progress

    - Closure governance and cross-workstream scope arbitration
      Owner: Avery Sinclair
      Status: In Progress
    
    - Extension authoring API completion pack and evidence validation
      Owners: Rohan Mehta, Grace Okafor
      Status: In Progress
    
    - UI scenario guide and rollout guidance for extension APIs
      Owners: Sofia Lindberg, Noah Patel
      Status: In Progress

    - Stabilization support for delivered chat/editor work
      Owners: Sofia Lindberg, Jonas Richter,
              Camila Torres, Victor Nguyen
      Status: In Progress

     **Metrics & Impact**
     - No quantitative delivery metrics were recorded in the recent diary entries.
     - Recent validation explicitly covered chat workflows, reasoning presentation, inline chat constraints, model management, accessibility-sensitive defaults, and adjacent editor improvements in one integrated pass on 2025-10-23.
     - The high-scale tool-selection refinement package addressed chat environments with more than 100 enabled tools and was handed forward with confusion-reduction and loading-state evaluation findings on 2025-10-22.
\end{flexprompt}
\captionof{figure}{Example of Project Status Question and Answer}
\label{fig:outputprojectstatusqa}
\end{center}

\begin{center}
\begin{flexprompt}[Example of Multi Hop People Question and Answer]{promptorange}{promptorangeborder}
"asked_on": "2025-10-31",
"asked_by": "Mei Foster",
"question": "Who owns the release narrative packaging work across the October framing flow?",
"answer": "Avery Sinclair, Ethan Wallace, and Mei Foster are the core owners, with distinct roles across the flow. Avery drove the kickoff, shaped and approved the theme and narrative wording, and led both integration and validation reviews. Ethan owned the decision-log and editorial-consistency side, carrying scope rules forward into the integrated package and claims review. Mei was the continuity owner across the work: she captured and normalized the baseline artifacts, turned scope constraints into working guidance, assembled and sent the handoff packages for the security and trust, editor experience, opening package, and integrated narrative, and built the final checklist and validated package used for sign-off."
\end{flexprompt}
\captionof{figure}{Example of Multi Hop People Question and Answer}
\label{fig:outputmultihopqa}
\end{center}

\begin{center}
\begin{flexprompt}[Example Decision and Feedback Recap Question and Answer]{promptorange}{promptorangeborder}
  "asked_on": "2025-07-28",
  "asked_by": "Rina Patel",
  "question": "What decisions and follow-ups came out of the latest reconciliation planning review for my workstream?",
  "answer": "The latest planning review locked the core reconciliation sequence and the operating rules around it. The team agreed that historical recomputation starts first, live processing stays on the old definition while new events are buffered, corrected history is applied through state patching, buffered events are replayed, and cutover happens only after that sequence is complete. They also finalized the customer-visible requirement that the Dashboard stays responsive during reconciliation and that status states clearly distinguish background processing from final cutover without showing mixed-definition analytics.\n\nOn the decision side, the group closed several open items. Readiness gating must be based on explicit completion evidence from historical outputs before patch-and-replay can start. Cutover cannot begin until corrected historical state has been applied and buffered events have been replayed without mixed-definition exposure. Cleanup of old-definition processing was explicitly moved to a controlled post-cutover step that only happens after successful health checks confirm no mixed-definition outputs remain visible.\n\nThe follow-up work from that review fed directly into workflow orchestration. I used the agreed stage sequence to define the reconciliation state machine, including submitted or accepted, running, ready-to-patch, patch-and-replay running, ready-to-cutover, completed, failed, and cleanup-pending states. Arjun was set to finish the coordinator rules for concurrent old-definition processing and buffered-event retention so the streaming control path can implement against a stable contract. Leah was assigned to complete the status propagation contract and failure semantics for serving and cutover consumers, including a distinct failed status with a retry-eligible flag so continuity handling stays stable while recovery remains possible."
\end{flexprompt}
\captionof{figure}{Example of Decision and Feedback Recap Question and Answer}
\label{fig:outputrecapqa}
\end{center}

% \begin{center}
% \begin{flexprompt}[QA]{promptorange}{promptorangeborder}
% Body
% \end{flexprompt}
% \captionof{figure}{Caption for orange prompt}
% \label{fig:prompt-orange2}
% \end{center}

\begin{center}
\begin{flexprompt}[Attribution]{promptorange}{promptorangeborder}
TASK:
You align generated Q&A items with the most relevant email messages so downstream systems can surface contextual replies.

GOAL:
For each question, determine which email thread and message best corresponds to the answer, justifying your choice using the email content and work log context. Flag any questions that cannot be mapped.

INSTRUCTIONS:
1. Review the email threads and understand their subject, participants, and discussion points.
2. Compare each question/answer pair with email content to find the closest semantic match.
3. Map each question to a single email (thread_id + email_id). If no suitable email exists, list the question id under unmapped_question_ids.
4. Provide a brief justification referencing specific snippets or themes from the chosen email.
5. Follow the attribution template exactly.

EVIDENCE QUALITY REQUIREMENTS (CRITICAL):
- The mapped email MUST contain SUBSTANTIVE EVIDENCE that directly supports the answer — meaning the email body includes the actual fact, outcome, decision, metric, or action described in the answer.
- DO NOT map to an email based solely on its subject line matching the topic of the question. The subject line alone is NOT sufficient evidence. The email body must contain the specific evidentiary content.
- DO NOT map to an email that only describes a precursor or preparatory step (e.g., "Let's schedule a review" or "I'll start looking into this") when the answer describes a result, outcome, or conclusion of that activity.
- When multiple emails touch the same topic, choose the one whose BODY contains the most specific and concrete evidence for the answer claims (e.g., specific metrics, decisions made, outcomes achieved).
- If only weak matches exist (e.g., same topic but no substantive overlap), list the question under unmapped_question_ids rather than forcing a poor mapping.
- In the justification, quote or closely paraphrase the specific sentences from the email BODY that support the answer. Do not justify by referencing only the subject line or generic topic alignment.

CONTEXT:

epic:
{epic}

Diary Context:
{week_log}

Email Threads:
{weekly_threads}

Q&A:
{weekly_qna}

Output Template:
{attribution_template}

OUTPUT:
- Return JSON matching the template with mappings and any unmapped ids.
- Use double quotes for all strings.
- Do not add explanations outside the JSON structure.

\end{flexprompt}
\captionof{figure}{Email QA Attribution Prompt}
\label{fig:attribuion}
\end{center}

% \begin{center}
% \begin{flexprompt}[Citations]{promptorange}{promptorangeborder}
% Body
% \end{flexprompt}
% \captionof{figure}{Caption for orange prompt}
% \label{fig:prompt-orange4}
% \end{center}

%%%%%%%%%%%%%%%%%%%%%%%%%%%%%%%%%%%%%%%%%%%%%%%%%%%%%%%%%%%%

\end{document}